\documentclass[sigplan,10pt,screen]{acmart}
\setcopyright{none}
\renewcommand\footnotetextcopyrightpermission[1]{} 

\AtBeginDocument{%
  }

\usepackage{cleveref}
\crefname{section}{§}{§§}
\Crefname{section}{§}{§§}

\usepackage{multirow}
\usepackage{makecell}
\usepackage[mathscr]{eucal}
\usepackage{subfigure}
\usepackage{pifont}
\usepackage[ruled,linesnumbered]{algorithm2e}
\usepackage{color}
\begin{document}

\title{PiPAD: Pipelined and Parallel Dynamic GNN Training on GPUs}

\author{Chunyang Wang}
\authornote{Both authors contributed equally to this work.}
\affiliation{%
	\institution{Beihang University}
	\city{Beijing}
	\country{China}
}
\email{wangchunyang@buaa.edu.cn}
\author{Desen Sun}
\authornotemark[1]
\affiliation{%
	\institution{Beihang University}
	\city{Beijing}
	\country{China}
}
\email{sy2006344@buaa.edu.cn}
\author{Yuebin Bai}
\affiliation{%
	\institution{Beihang University}
	\city{Beijing}
	\country{China}
}
\email{byb@buaa.edu.cn}

\begin{abstract}
\footnote{To appear at PPoPP'23.} Dynamic Graph Neural Networks (DGNNs) have been broadly applied in various real-life applications, such as link prediction and pandemic forecast, to capture both static structural information and temporal characteristics from dynamic graphs. Combining both time-dependent and -independent components, DGNNs manifest substantial parallel computation and data reuse potentials, but suffer from severe memory access inefficiency and data transfer overhead under the canonical one-graph-at-a-time training pattern. To tackle the challenges, we propose PiPAD, a \underline{\textbf{Pi}}pelined and \underline{\textbf{PA}}rallel \underline{\textbf{D}}GNN training framework for the end-to-end performance optimization on GPUs. From both the algorithm and runtime level, PiPAD holistically reconstructs the overall training paradigm from the data organization to computation manner. Capable of processing multiple graph snapshots in parallel, PiPAD eliminates the unnecessary data transmission and alleviates memory access inefficiency to improve the overall performance. Our evaluation across various datasets shows PiPAD achieves $1.22\times-9.57\times$ speedup over the state-of-the-art DGNN frameworks on three representative models.
\end{abstract}



\maketitle
\pagestyle{plain}
\section{Introduction}
Graph Neural Networks (GNNs) have been broadly adopted to process graph-structured data and extract the underlying dependencies \cite{abadal2021computing} in varieties of graph-related applications ranging from node classification \cite{garcia2017learning}, recommendation \cite{ma2020memory} to link prediction \cite{v2018graph}. Among numerous variants of GNNs, such as Graph Convolution Network (GCN) \cite{welling2016semi}, a common and prevalent method is combining both graph (aggregation) and neural (update) operations. The aggregation function collects and aggregates feature vectors from the neighbors of each node while the update phase utilizes neural network operations, such as a fully connected (FC) layer, to update each vertex with the aggregated information. With retrieving the sparse adjacent matrices that renders irregular memory accesses and sparse computation, the Sparse-Dense Matrix-Matrix-Multiplication-like (SpMM-like) aggregation operation \cite{huang2020ge} is generally considered as the main bottleneck of GNN and attracts lots of research attentions \cite{huang2020ge, wang2021gnnadvisor, huang2021understanding, fu2022tlpgnn}.

In many real-world scenarios, such as financial transaction \cite{chakaravarthy2021efficient}, social media \cite{li2021cache} and molecular biology \cite{fout2017protein}, the topology and node features of graphs may dynamically evolve over time. According to the partition principle and granularity, the dynamic graph representation can be categorized into two classes \cite{kazemi2020representation, chakaravarthy2021efficient, zhou2022tgl}: Continuous Time Dynamic Graphs (CTDGs) and Discrete Time Dynamic Graphs (DTDGs). And the GNN research community proposes various dynamic GNNs (DGNNs) to capture both temporal and structural information in the mobility graph. For DTDGs that represent the dynamic graph as a sequence of \textit{snapshots} sampled at regular intervals, a general method is to use static GNNs (e.g., GCN) for spatial graph learning on individual snapshots at all timesteps while deploying Recurrent Neural Networks (RNNs) to obtain temporal characteristics among different snapshots \cite{chakaravarthy2021efficient, guan2022dynagraph, skarding2021foundations}. Besides, the \textit{sliding window} mechanism that feeds multiple continuous snapshots to the model simultaneously, is widely adopted to better capture the temporal dependence and improve the accuracy \cite{skarding2021foundations, pareja2020evolvegcn, panagopoulos2021transfer, guan2022dynagraph}. In this paper, we focus on \textbf{DTDG-based} DGNNs and refer to the sliding window as \textit{frame} for simplicity.

Compared to traditional GNNs, DGNNs integrate time-series components operating a mass of snapshots, which leads to \textbf{two main performance issues}. First, since DGNN training requires updating graph snapshots continuously along the timeline, the data transfer overhead dominates overall training time and further exacerbates GPU underutilization that already exists due to the memory-intensive GNN aggregation operation. Our preliminary experiments (\cref{sec:motivation1}) based on the state-of-the-art DGNN framework, PyTorch Geometric Temporal (PyGT) \cite{rozemberczki2021pytorch}, show the data transmission via PCIe occupies nearly $39\%$ of total execution and GPU utilization is lower than $42\%$ on average. Second, as introduced detailedly in \cref{sec:motivation0}, the combination of time-independent (GNN) and time-dependent (RNN) components produces substantial parallelism opportunities (e.g., executing those independent operations from multiple snapshots concurrently) while the snapshot overlap among adjacent frames offers plenty of data reuse chances. These acceleration potentials desperately need to be sufficiently exploited. 

Over the most recent years, certain research efforts have been made to realize efficient DGNN computation on GPUs \cite{rozemberczki2021pytorch, zhou2022tgl, chakaravarthy2021efficient, li2021cache, guan2022dynagraph} and some of them involve the above issues. However, the related solutions either mainly focus on the scaling of large graphs and distributed training \cite{chakaravarthy2021efficient} or restrict their input scenarios by assuming the graph topology (or node features) not changing along the time \cite{li2021cache, guan2022dynagraph}, which weakens the applicability. Furthermore, they all fail to tackle both aforementioned issues at the same time for maximal end-to-end training performance optimizations.

Instead of analyzing the two problems separately, we address them from a more holistic angle and in a collaborative way. The idea is motivated by two key discoveries. First, real-world dynamic graphs normally change at a slow pace rendering massive topology overlaps among the adjacent snapshots \cite{chakaravarthy2021efficient}. Second, except the irregular access pattern of the SpMM-like aggregation, GNN suffers from other kinds of memory inefficiency stemming from the diverse node feature dimensions. Performing the aggregation over single graph tends to incur the \textit{bandwidth unsaturation}, \textit{low thread utilization} or \textit{request burst} problems (\cref{sec:motivation2}). In summary, the canonical one-snapshot-at-a-time training paradigm inevitably causes not only substantial redundant data transmission but also inefficient memory accesses. Therefore, our insight is that the GPU underutilization and parallelism neglect issues could be alleviated simultaneously. Specifically, the expensive aggregation operates on the adjacent matrices that are exactly the main source of redundant data transfer. Extracting the overlapped topology and conducting single time-irrelevant aggregation operation for multiple snapshots can not only reduce the communication volume but also have more opportunities to enable coalescent memory accesses and address the low bandwidth utilization issue.

\begin{figure*}[htbp]
	\centering
	\begin{minipage}[b]{.26\linewidth}
		\centering
		\includegraphics[width=\linewidth]{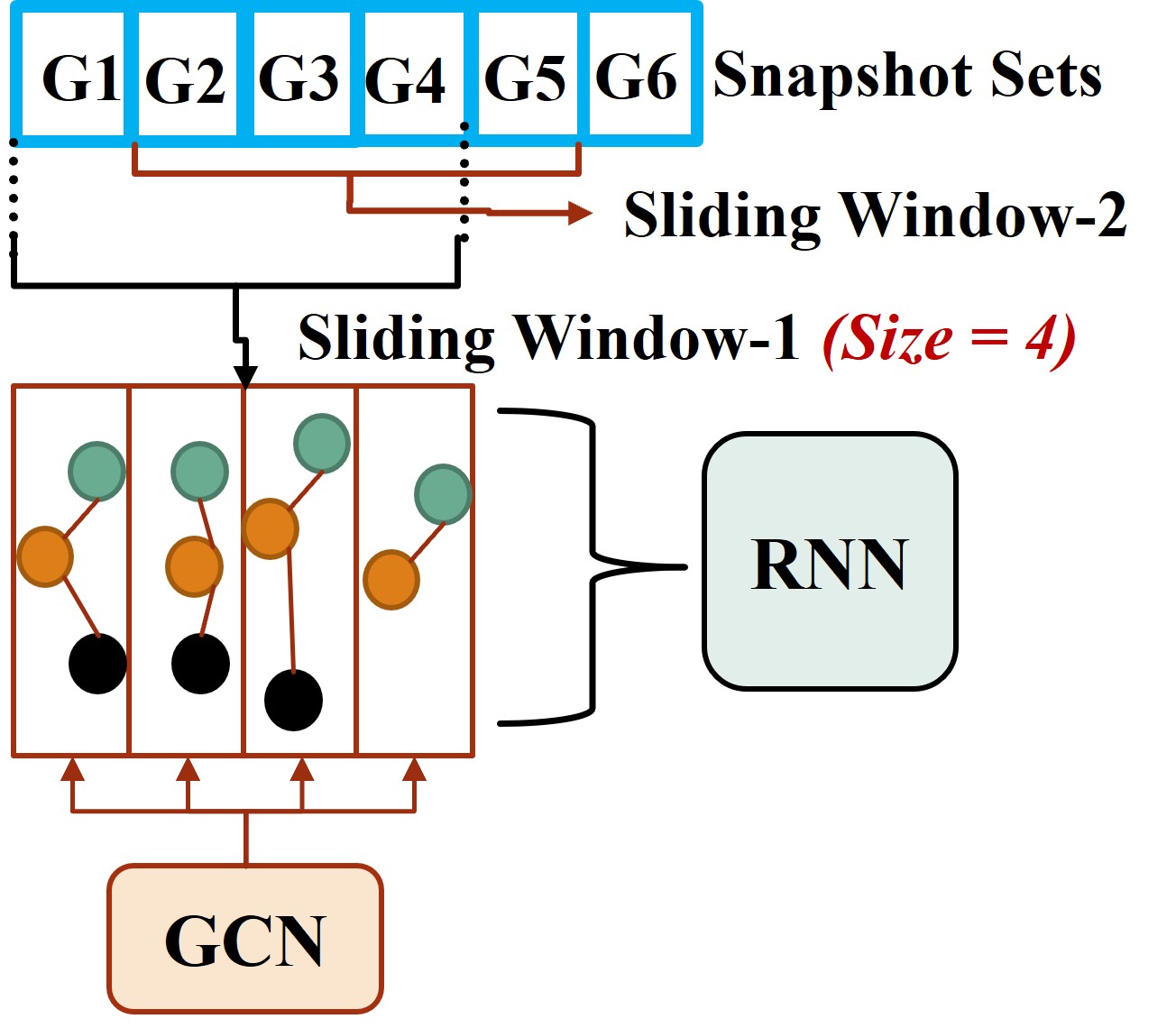}
		\caption{DGNN Execution Flow.}
		\label{fig:sliding}
	\end{minipage}
	\hspace{.1in}
	\begin{minipage}[b]{.66\linewidth}
		\centering
		\subfigure[MPNN-LSTM]{ 
			\centering    
			\includegraphics[width=.24\linewidth]{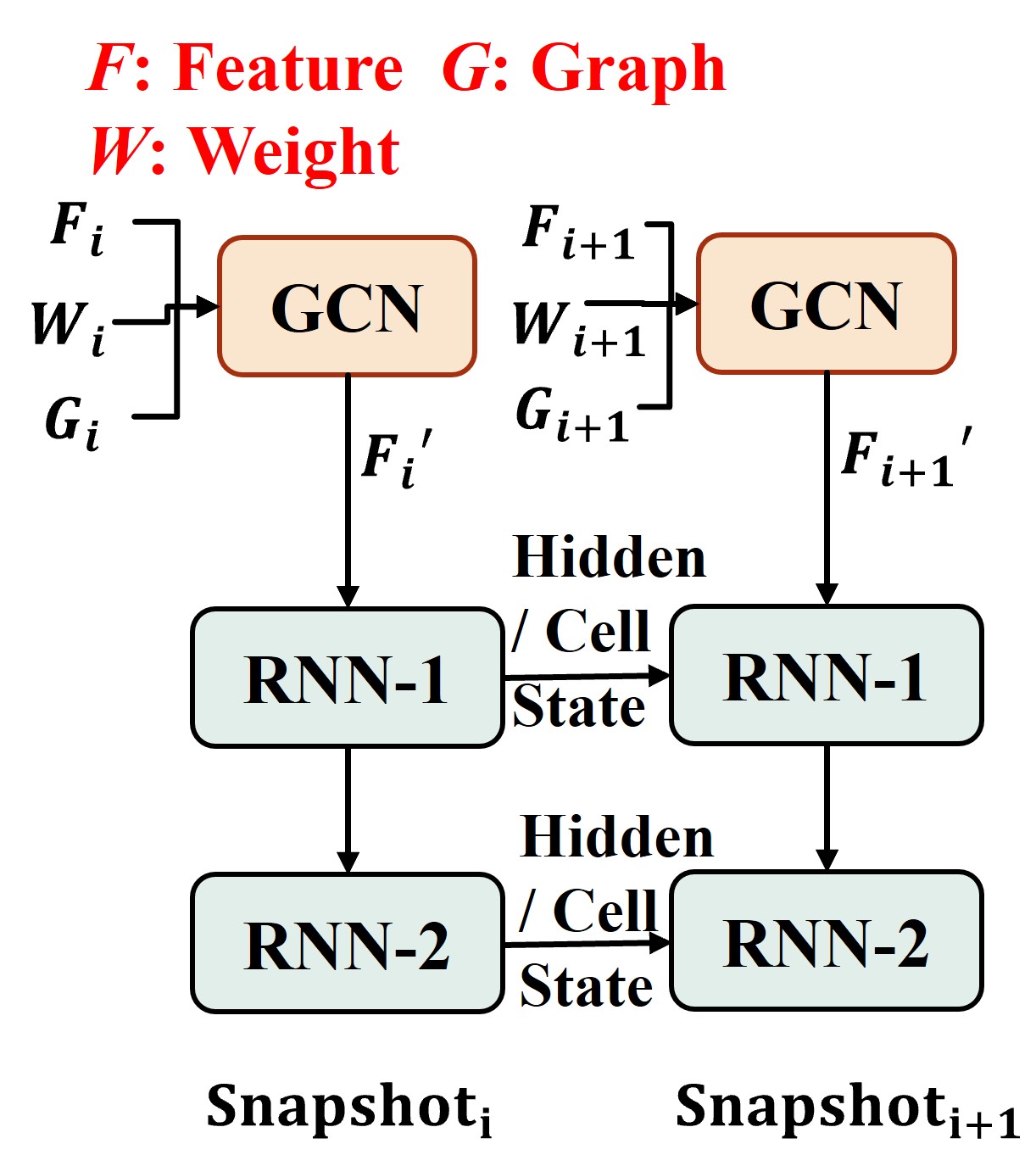}
			\label{fig:MPNN-LSTM}
		}
		\subfigure[EvolveGCN]{   
			\centering    
			\includegraphics[width=.3\linewidth]{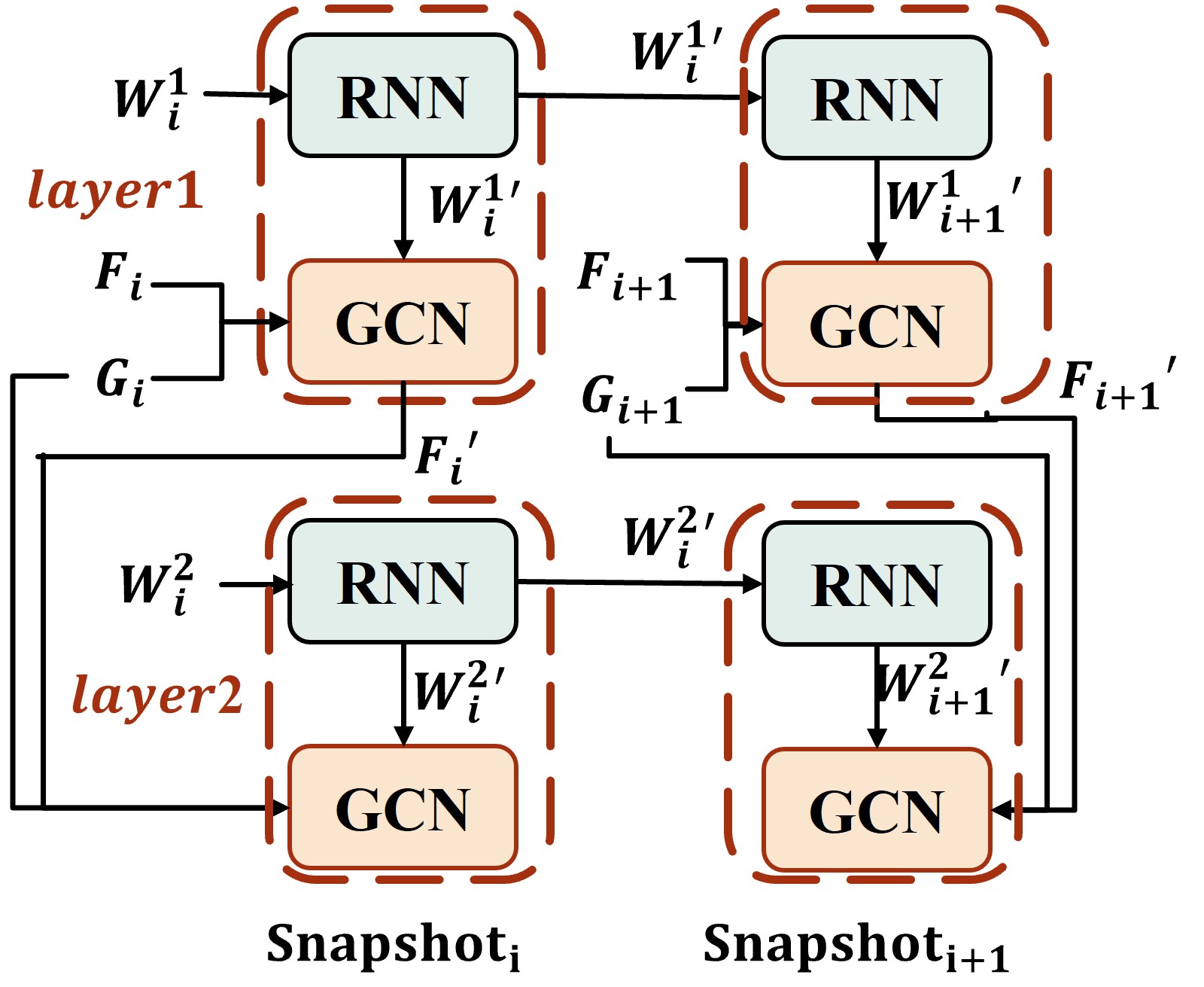}  
			\label{fig:EvolveGCN}
		}
		\subfigure[T-GCN]{   
			\centering    
			\includegraphics[width=.36\linewidth]{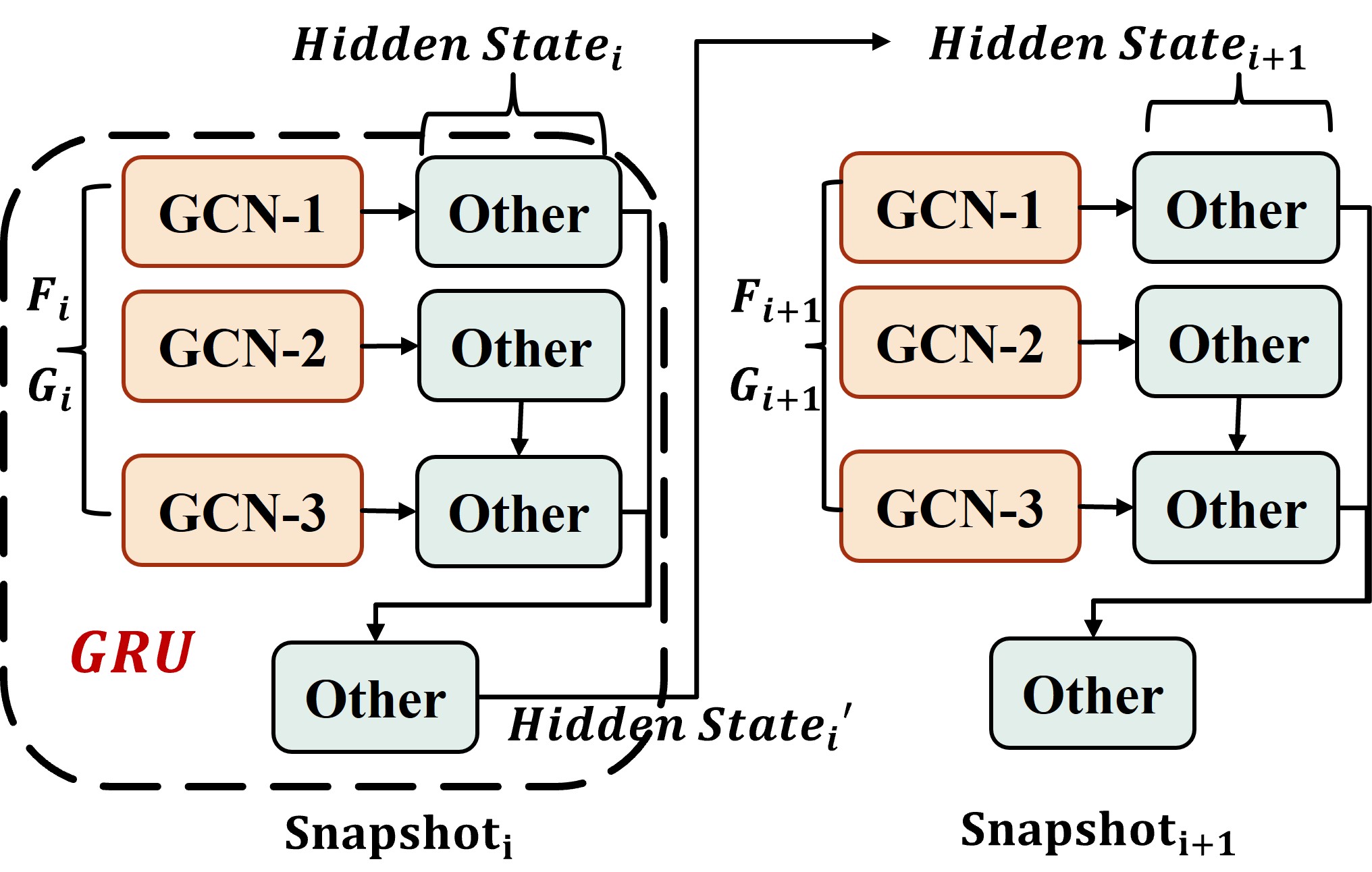}  
			\label{fig:T-GCN}
		}
		\caption{DGNN Model Structure. $\rightarrow$: Data Propagation.}
		\label{fig:model}
	\end{minipage}
\end{figure*}

To this end, we propose PiPAD, a \underline{\textbf{Pi}}pelined and \underline{\textbf{PA}}rallel \underline{\textbf{D}}GNN training framework to optimize the end-to-end performance on single GPU from both algorithm and runtime level. Without making any assumption on the input scenarios, PiPAD enables pipelined and parallel execution via reconstructing the overall DGNN training process from the data organization and transfer method to computation manner. \textbf{First}, capitalizing on the gradual changing characteristic of dynamic graphs, we propose a slice-based graph representation format and overlap-aware multi-snapshot transfer method. The design objective is to reduce the data loading time from CPU to GPU and facilitate the efficient parallel computation. \textbf{Second}, we devise the intra-frame parallelism and inter-frame reuse mechanism to sufficiently unleash the acceleration potential in DGNN. The former realizes a dimension-aware parallel GNN with thread-aware slice coalescing to boost the aggregation and locality-optimized weight reuse for the update. Our parallel GNN supports processing multiple snapshots simultaneously for better computation performance and alleviating the memory access inefficiency. Inter-frame reuse can eliminate the redundant transfer and computation overhead to the maximum extent. \textbf{Third}, to optimize the overall training performance, we implement a pipeline execution framework to orchestrate CPU-side operations, data transmission over PCIe and GPU computation. Besides, PiPAD integrates a runtime tuner to dynamically adjust the parallelism- and reuse-level of DGNN training for better adaptation to various datasets.

Since our objective is to reduce the end-to-end execution time, we conduct our additional data processing (graph slicing) online and avoid any offline model-scale profiling. Specifically, PiPAD only previously profiles and analyzes our parallel GNN kernel to guide the online tuning. As GCN is the most popular GNN employed in DGNNs nowadays \cite{skarding2021foundations}, this paper targets GCN-based models. But with the SpMM-like aggregation being the foundation of mainstream GNNs \cite{huang2020ge, huang2021understanding} (e.g., Graph Attention Network \cite{v2018graph}), our methodology thus can be applied to various types of DGNNs.

Overall, this paper makes three contributions:
\begin{itemize}
	\item {} We conduct a comprehensive analysis of the execution and performance characteristics for three typical DGNN models (\cref{sec:dgnn} \& \cref{sec:motivation}). Several key discoveries of DGNN training are revealed and motivate our design.
\end{itemize}
\begin{itemize}
	\item {} We propose PiPAD to optimize the end-to-end performance for single-GPU DGNN training. Incorporating the overlap-aware data organization (\cref{sec:transfer}), intra-frame parallelism (\cref{sec:parallel}), pipeline execution mechanism (\cref{sec:Pipe}) and inter-frame reuse with dynamic tuning technique (\cref{sec:reuse}), PiPAD holistically enables pipelined and parallel training in a multi-snapshot manner.
\end{itemize}
\begin{itemize}
	\item {} Our evaluation across various datasets shows PiPAD achieves $1.22\times-9.57\times$ speedup over the state-of-the-art DGNN frameworks on three representative models.
\end{itemize}

\section{Background}
\subsection{Dynamic Graph Neural Networks}\label{sec:dgnn}
A DTDG is a ordered set of snapshots $\{\mathcal{G}^1, \mathcal{G}^2 \cdots \mathcal{G}^t\}$, where $\mathcal{G}^t=\{\mathcal{V}^t, \mathcal{E}^t\}$ represents the snapshot with vertices $\mathcal{V}^t$ and edges $\mathcal{E}^t$ at the timestep $t$. For one individual snapshot $\mathcal{G}$ and a GNN with K layers, let $h^{k-1}_v$ denotes the feature vector of vertex $v$ at layer $k-1$ and $\mathcal{N}(v)$ denotes all one-hop neighbors of vertex $v$. The computing pattern of GNN on layer $k$ can be represented in Equation \ref{eq1}. After K layers of propagation, we will get the final features for each vertex. \begin{equation}\label{eq1}h^{k}_v = \textbf{Update}^k(\textbf{Aggregation}^k(h^{k-1}_u | {\forall}u \in \mathcal{N}(v) \cup \{v\}))\end{equation}

Different GNNs differ in the specific functions used for the two key phases. In GCN, the aggregation processes the gathered features with \textit{mean} function to obtain the aggregation results $a^{k}_v$ (namely hidden representations) for each vertex while the FC-layer-based update computes $h^{k}_v$ using the weights $w^k$ and bias parameters $b^k$.

DTDG-based DGNNs follow the sliding-window/frame processing pattern and may encompass multiple layers. As shown in Figure \ref{fig:sliding}, inside each layer, GNN components learn the structural information from each snapshot while RNN modules capture the temporal characteristics along the timeline. DGNNs can be classified into two categories \cite{skarding2021foundations}: stacked and integrated. We choose three representative models for the analysis and evaluation in the rest of this paper.

\textbf{MPNN-LSTM} \cite{panagopoulos2021transfer} (Figure \ref{fig:MPNN-LSTM}) stacks one 2-layer GCN and two LSTM \cite{hochreiter1997long} models in a plain manner. LSTM operations are applied over the features produced by GCN. The only data dependence across multiple snapshots exists in the hidden state updating process for LSTM.

\textbf{EvolveGCN} \cite{pareja2020evolvegcn} (Figure \ref{fig:EvolveGCN}) involves a 1-layer GCN and one GRU \cite{cho2014learning} in each of its two layers. The GCN in the second layer takes outputs from the first layer as input features. Different from MPNN-LSTM, EvolveGCN enables the weights evolving by applying the RNN module over GCN weight matrices, which generates cross-snapshot dependence. The survey \cite{skarding2021foundations} categories EvolveGCN into the integrated type.

\textbf{T-GCN} \cite{zhao2019t} (Figure \ref{fig:T-GCN}) integrates several 1-layer GCNs into GRU by replacing the original general matrix-matrix multiplication (GEMM). The hidden state propagation over the timeline creates the dependence similar to MPNN-LSTM.

\subsection{GNN Acceleration}
\textbf{Static GNN.} Over the last few years, there have been substantial research achievements for static GNN acceleration on GPUs covering general runtime frameworks \cite{fey2019fast, wang2019deep, ma2019neugraph, zhu2019aligraph}, the SpMM-like aggregation optimization \cite{huang2020ge, wang2021gnnadvisor, huang2021understanding, fu2022tlpgnn} and the scaling of distributed training \cite{yang2022gnnlab, wang2021flexgraph, thorpe2021dorylus, li2022hyperscale, su2021adaptive, wan2022pipegcn, kaler2022accelerating}.

\textbf{DGNN.} PyGT \cite{rozemberczki2021pytorch} and TGL \cite{zhou2022tgl} are two general frameworks aiming to implement the ubiquitous support for as many DGNN models as possible. By contrast, DynaGraph \cite{guan2022dynagraph} and CacheG \cite{li2021cache} focus on optimizing a certain group of models. Also discovering the parallelism/reuse opportunities (\cref{sec:motivation0}) in DGNN, they leverage the timestep fusion and intermediate results cache to improve the performance, respectively. However, DynaGraph only targets integrated DGNNs and assumes the graph topology not evolving over time while CacheG demands the node features remaining unchanged. Cambricon-G \cite{song2021cambricon} and ESDG \cite{chakaravarthy2021efficient} both involve DGNN's data transfer overhead issue (\cref{sec:motivation1}). The hardware accelerator solution Cambricon-G, devises a cuboid-based processing architecture that supports the fine-grained data transmission to avoid unnecessary snapshot topology updating. ESDG employs a similar graph-difference based transfer method to reduce the communication volume for the scaling of large graphs and multi-GPU training. But those two studies neglect the intrinsic parallelism potentials in DGNN and blunder away the chance of fulfilling further acceleration.

\textbf{Pipeline/parallel training} for GNN or traditional DNN is widely adopted in the distributed environment \cite{yang2022gnnlab, wan2022pipegcn, su2021adaptive, narayanan2019pipedream, huang2019gpipe, narayanan2021memory} and concurrent multi-model computation \cite{bai2020pipeswitch, sun2022cognn}. OOB \cite{oh2022out} and nimble \cite{kwon2020nimble} are two similar studies to us enabling parallel training for single DNN on single GPU. But the former's motivation lies on the backward propagation while the latter focuses on using CUDA Graph \cite{cudagraph} and dependency pre-analysis mechanisms to remove the multi-stream overhead. Their work are orthogonal to our concern.

\textbf{Summary.} Works about the SpMM-like aggregation optimization for static GNNs are more general and can be applied to the DGNN domain. But some studies \cite{huang2020ge, wang2021gnnadvisor} rely on the onerous node reordering (up to seconds per snapshot) that could badly damage the end-to-end performance. Besides, these solutions and the existing DGNN optimization methods cannot address both key issues simultaneously.

\section{Motivation}\label{sec:motivation}
This section first reveals the bottleneck and underlying execution inefficiency of DGNN training via preliminary experiments with our evaluation setting (\cref{sec:setup}). Then we dissect the data reuse and parallel computation potentials, which can be leveraged to resolve these issues simultaneously.
\begin{figure*}[htbp]
	\centering
	\begin{minipage}[t]{.31\linewidth}
		\centering
		\includegraphics[width=\linewidth]{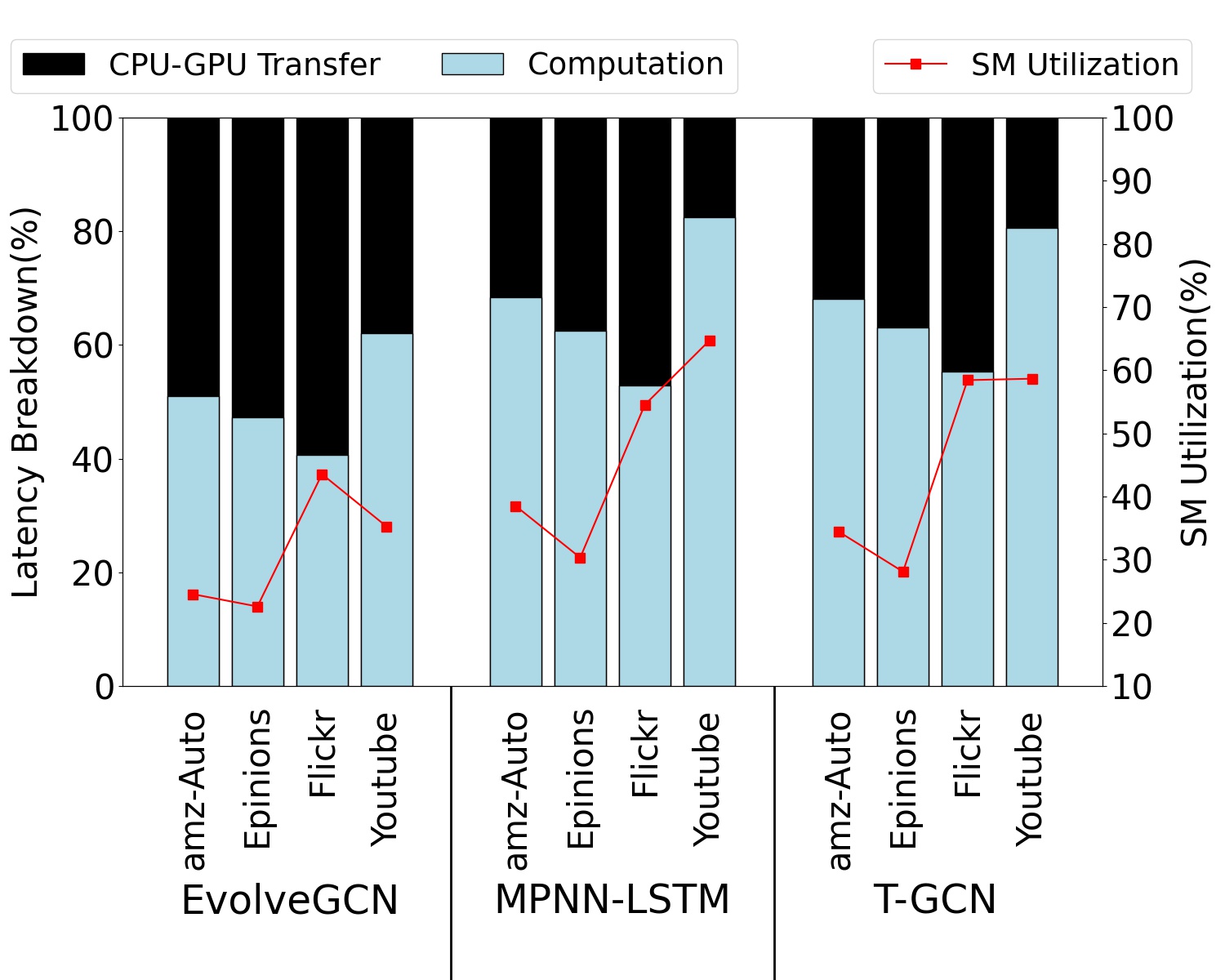}
		\caption{Latency Breakdown and SM Utilization of DGNN Training.}
		\label{fig:breakdown}
	\end{minipage}
	\hspace{.1in}
	\begin{minipage}[t]{.31\linewidth}
		\centering
		\includegraphics[width=\linewidth]{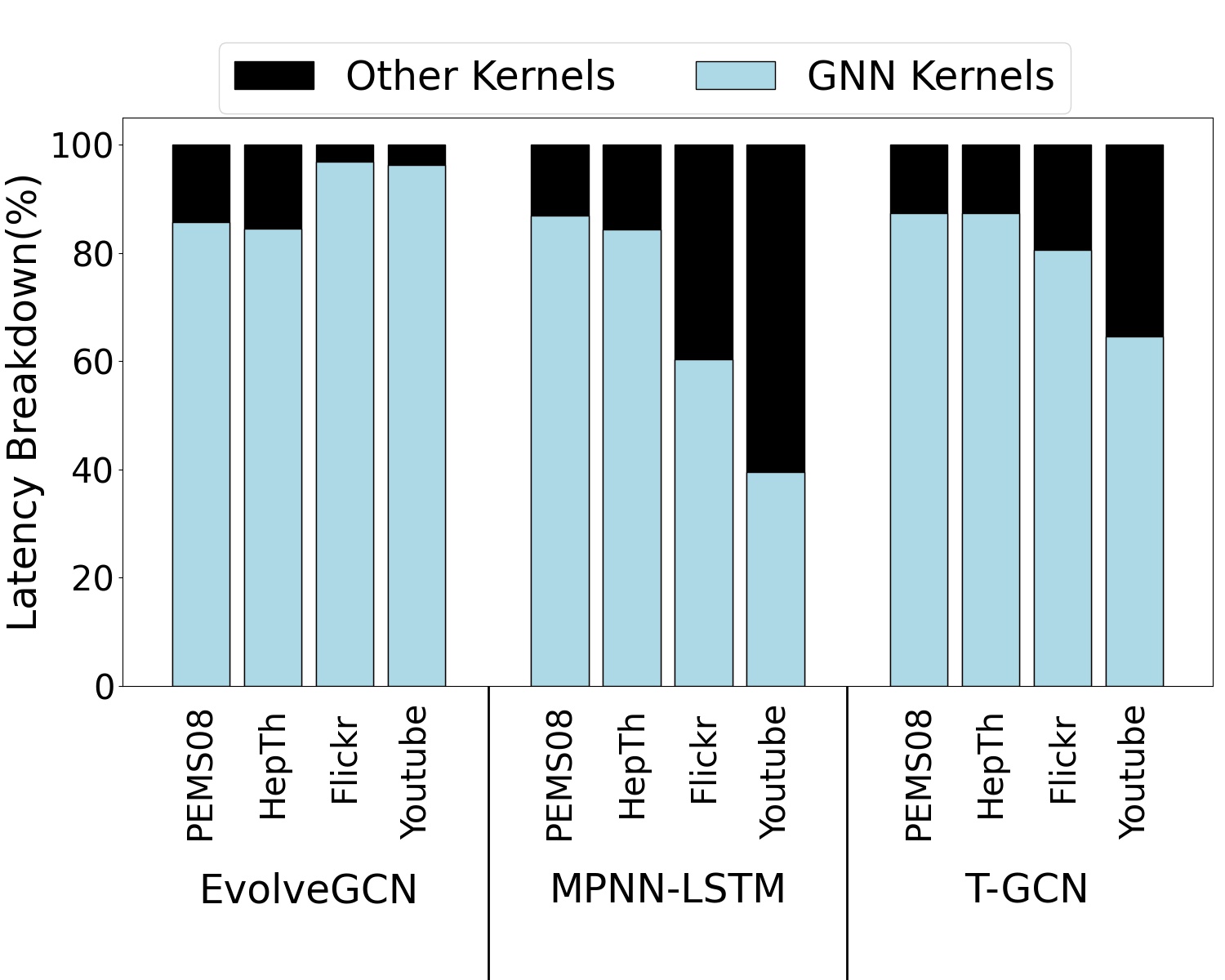}
		\caption{Breakdown of GPU Computation Time in DGNN Training.}
		\label{fig:GCNbreakdown}
	\end{minipage}
	\hspace{.1in}
	\begin{minipage}[t]{.31\linewidth}
		\centering
		\includegraphics[width=\linewidth]{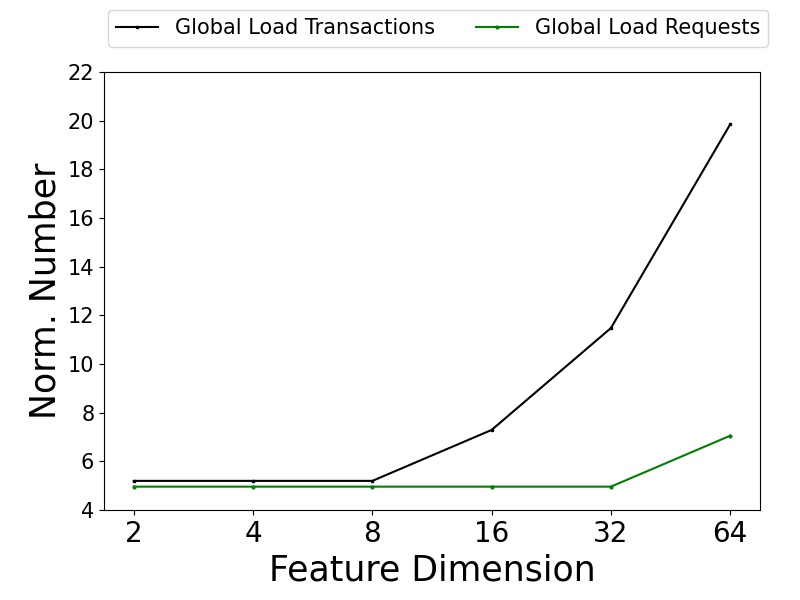}
		\caption{The number of Global Memory Requests and Transactions.}
		\label{fig:transaction}
	\end{minipage}
\end{figure*}


\subsection{Performance Bottlenecks of DGNN Training}\label{sec:motivation1}
\textbf{Data transfer overhead}. We analyze the overall performance characteristic of DGNN training using PyGT, which is built on top of the popular GNN computing framework PyTorch Geometric (PyG) \cite{fey2019fast}. Compared to the static GNN targeting single graph, DGNN operates on a snapshot sequence and needs to constantly load new graph data from CPU to GPU. Figure \ref{fig:breakdown} depicts the breakdown of GPU-related training time (left-axis) and SM utilization (right-axis). As presented, the arduous data transfer occupies the total execution time with an average $38.7\%$ proportion. Subsequently, this leads to severe GPU underutilization where Streaming Multiprocessor (SM) utilization measured by PyTorch Profiler \cite{profiler} is lower than $41.2\%$ on average. Leveraging CUDA streams to enable asynchronous data transmission is a straightforward and elementary way to ease the problem.

\textbf{Topology overlap}. The changing rate of the topology among adjacent snapshots in real-life dynamic graphs is generally limited (nearly $10\%$ on average across our datasets). ESDG \cite{chakaravarthy2021efficient} utilizes this fact to employ a graph-difference transfer method that only updates changed parts of the topology, but still follows the one-snapshot-at-a-time training manner.

\subsection{Memory Access Inefficiency in GNN}\label{sec:motivation2}
Still being the major computation burden in DGNN models (Figure \ref{fig:GCNbreakdown}), GNN suffers from the low memory access efficiency. The access irregularity in the aggregation, brought by the sparsity of input graphs, is revealed by previous GNN studies \cite{huang2020ge, wang2021gnnadvisor, huang2021understanding, fu2022tlpgnn}. They refactor access patterns to the sparse adjacent matrices and accordingly implement the locality-aware computation to increase the parallelism. However, there are other types of inefficiency stemming from the dimension of dense matrices (input or aggregated feature vectors), neglected by those existing solutions. In a common implementation of SpMM, one warp takes charge of processing single element from the sparse matrix and one entire row from the dense matrix for each outer iteration. To leverage the locality and reduce expensive off-chip accesses, GPU cores load the features from global memory and store the intermediate results of the current iteration on shared memory for reuse in the next iteration. Note that mainstream GPUs (e.g., NVIDIA) normally employ a minimum 32-byte granularity for global memory access. Besides, following the Single Instruction Multiple Thread (SIMT) fashion, a warp consisting of 32 threads can fetch 128 ($32\times4$) bytes at most with one request. Considering the row length in bytes of the feature matrix equal to 4 times of its feature dimension (\textit{F}), there are two types of memory access inefficiency:
\begin{itemize}
	\item \textbf{Bandwidth unsaturation} with \textit{F} less than 32/4, which means that the size of useful data is smaller than the minimum access granularity of single transaction.
	\item \textbf{Request burst} with \textit{F} larger than 128/4, which means that a warp accessing one row needs multiple requests to both shared and global memory to finish.
\end{itemize}

We further demonstrate the above insights via experiments of running GCN with GNNAdvisor \cite{wang2021gnnadvisor}, a cutting-edge GNN aggregation optimization work. As shown in Figure \ref{fig:transaction}, when the feature length lower than 8, the number of global memory requests ($\#R$) and transactions ($\#T$) are nearly the same and both barely change. Then $\#T$ increases as the dimension exceeds 8 while $\#R$ begins to rise when the dimension larger than 32. Considering the diversity of dimension features used in real-life graph-based applications, these issues definitely need to be resolved. Moreover, assigning one row to one warp when processing the feature matrix with F less than 32, means many threads inside the warp stay idle during the execution. The \textbf{low thread utilization} issue can be reflected by the \textit{warp\_execution\_efficiency} metric (\cref{sec:detailed}). 
\subsection{Opportunities for Data Reuse and Parallelism}\label{sec:motivation0}
\textbf{Data Reuse}. The forward stride size of the frame mechanism (Figure \ref{fig:sliding}) is normally set to 1 for sufficient temporal interaction capture among all snapshots. Therefore, there are plenty of overlaps among contiguous frames, which means some related redundant data transfer and computation can be avoided. Specifically, the aggregation operation in the first GCN layer operates over the adjacent matrix of the graph topology and the original node features, which is independent of the parameter updating along the timeline. We can cache the related aggregation results ($a^{k}_v$ in \cref{sec:dgnn}) and reuse them in the next frame or training epoch.

\begin{figure*}
	\centering
	\includegraphics[scale=0.31]{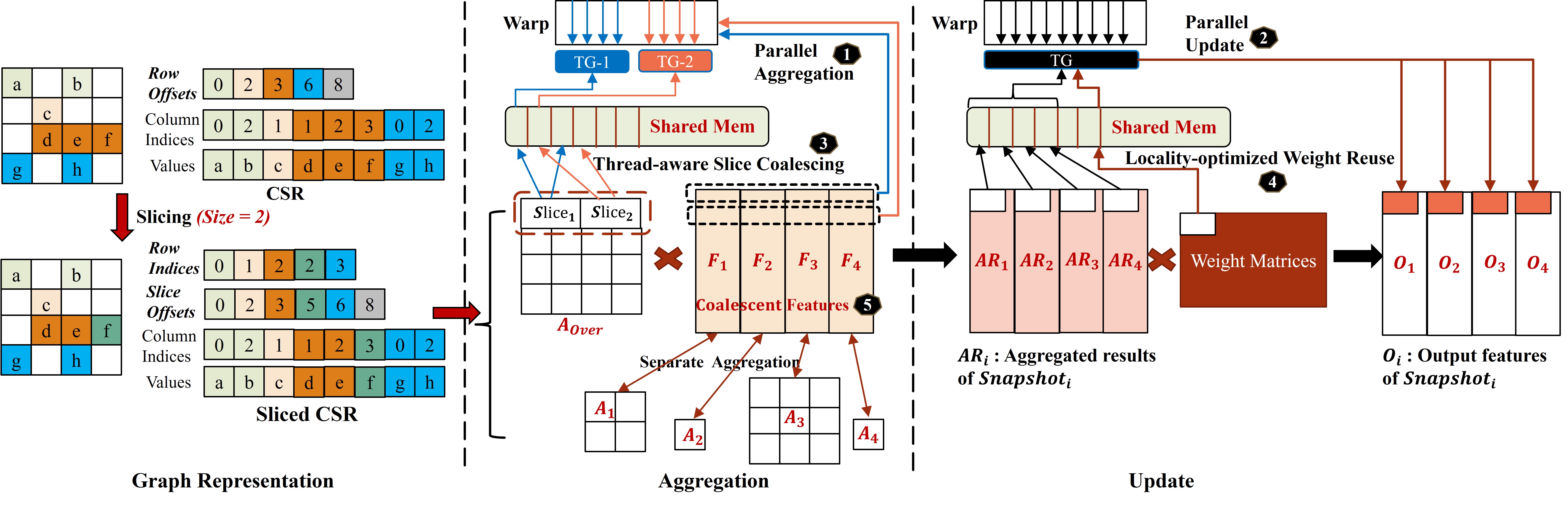}
	\caption{\label{fig:computation} An Example of the Sliced Graph Representation and the Parallel GNN Computation. $F_i$: Input Features of $Snapshot_i$. $A_i$: Adjacent Matrix of the Exclusive Parts of $Snapshot_i$. $A_{Over}$: Adjacent Matrix of the Overlap Part. TG: Thread Group.}
\end{figure*}
\textbf{Parallelism}. As depicted in Figure \ref{fig:model}, the cross-snapshot dependence of three various DGNNs all exists in the time-dependent RNN components. It means that the same operation of GNN for different snapshots can be conducted in parallel for better performance. With careful design and optimizations, the multi-snapshot processing manner may hit two birds with one stone. Specifically, we first take \textit{topology overlap} into consideration cooperatively and design a new graph format that can extract the overlaps efficiently to eliminate the redundant transfer. Then we \textbf{devise a ``parallel'' GNN computation pattern that can process multiple graphs simultaneously. The basic idea is to directly perform the aggregation on the overlap topology of one snapshot group with all their feature matrices}. In this way, we can enable coalesced memory accesses to multiple features and alleviate the bandwidth unsaturation issue. Further, \textbf{employing flexible access granularities for huge dimension situations can avoid the request burst}.


\section{PiPAD}\label{sec:PiPAD}
We holistically design and implement PiPAD to facilitate the end-to-end DGNN training performance. Our optimization is three-fold: reducing data transfer volume via the overlap-aware data organization and inter-frame reuse, accelerating GNN computation with intra-frame parallelism and maximizing the overlap among different operations through the pipeline. This section presents those mechanisms in detail.

\subsection{Overlap-aware Data Organization}\label{sec:transfer}
By extracting the overlap topology among adjacent snapshots and constructing a new individual adjacent matrix for it, we can avoid the needless transmission and further realize the parallel computation idea. Due to the frame mechanism, one snapshot may be coalesced with separate snapshots for processing in different frames. Therefore, the overlap analysis and extraction will be conducted constantly throughout the opening training epochs. The widely-used format to store the sparse adjacent matrix, Compressed Sparse Row (CSR) manages the graph using a coarse row granularity and a tightly ordered layout (Figure \ref{fig:computation}). These characteristics directly restrict the flexibility of overlap extraction leading to poor efficiency. Besides, the sparsity in real-world graphs could easily cause load imbalance in SpMM computation since a warp is normally set to handle one or several rows of the adjacent matrix based on the CSR representation way.

\textbf{Slice-based graph representation.} There are two choices to support a finer storing granularity and tackle the above challenges: two-dimensional \textit{tile} and one-dimensional \textit{slice}. The former (e.g., Block CSR \cite{blockCSR}) splits the original matrix into multiple smaller matrices while the latter divides each row and transforms the whole matrix into many slices. Since the non-zero elements (nnz) in each tiled sub-matrix are discrete leading to the more uneven sparsity, the slice way not only incurs less space overhead but also offers better load balance with the compression. Therefore, we devise a sliced CSR format shown in Figure \ref{fig:computation}. As single slice is within one row, we modify the original \textit{Row Offsets} array in CSR to \textit{Row Indices} (RI) and add a new array \textit{Slice Offsets} (SO). RI indicates the row index of every slice and SO stores the offset of the first element of each slice in \textit{Column Indices} field. Our method manages the whole graph at a slice granularity and each slice stores certain nnz with a upper bound (2 in Figure \ref{fig:computation}). This can enable quick slice-grained overlap construction and assist to achieve better load balance in the computation.

\textbf{Overlap-aware data transfer.} Introduced later, PiPAD performs the parallel training by dividing each frame into several \textit{partitions} that each contain continuous snapshots. We thus adjust the data transfer into a partition-grained pattern. The topology data (adjacent matrices) are regrouped as one overlap part for all snapshots and several exclusive parts for each to reduce the transmission volume. Then we utilize separate GPU streams and pinned memory to enable asynchronous data transfer in the on-demand manner.

\textbf{Space overhead.} The main cost comes from the modified RI and newly added SO. Let $\#nnz$ denotes the number of nnz in the matrix. Our sliced CSR requires $(2\times \#nnz + 2\times\#Slice + 1)$ space while CSR needs $(2\times \#nnz + \#vertices + 1)$. Another popular graph representation way Coordinate Format (COO), employed by PyG, uses three arrays to respectively store all nnz's values, the row and column indices with the $(3\times \#nnz)$ space usage. We set single slice can hold up to 32 nnz and our spatial cost thus normally falls in between CSR and COO.

\subsection{Intra-frame Parallelism}\label{sec:parallel}
We implement intra-frame parallelism based on our \textit{dimension-aware parallel GNN} that performs the aggregation and update operations over multiple snapshots simultaneously (\ding{182} and \ding{183} in Figure \ref{fig:computation}). We further devise three key optimizations: vector-memory-instruction based memory access for the large-dimension situation, \textit{thread-aware slice coalescing} (\ding{184}) to resolve the low thread utilization issue for the small-dimension case and \textit{Locality-optimized weight reuse} (\ding{185}) to enable the efficient parallel update.

\textbf{Dimension-aware parallel GNN}. As analyzed in \cref{sec:motivation0}, two types of memory access inefficiency can be alleviated through the multi-snapshot computation manner and flexible access granularities. First, coalescing the feature matrices from different snapshots to perform one aggregation can more efficiently utilize the bandwidth especially for those graphs with node features in the small dimension. Following the overlap-aware data organization, we split the original aggregation into two parts: the parallel aggregation on adjacent matrices of the overlap with all features of one snapshot group (referred to as the \textit{coalescent} features: \ding{186} in Figure \ref{fig:computation}), and the non-parallel aggregations on the exclusive topology parts with respective features from each snapshot. Second, leveraging the vector memory instructions \cite{vectormem} that support loading 32/64/128 floats for one request is a feasible option for the request burst problem. Based on those vector instructions with different data-fetching widths, we realize loading the data in an efficient larger-grained way when processing graphs with features in the large dimension (larger than 32). We introduce our aggregation implementation with \textit{slice coalescing} together later.

\begin{algorithm} 
	\caption{Parallel aggregation \& Slice coalescing}
	\label{alg:1}
	\LinesNumbered 
	\footnotesize
	\KwIn{features, slice\_size, coalesce\_num}
	\KwOut{aggregation\_result}
	
	\scriptsize
	$lane\_id = \textbf{getIndexInWarp}()$ \\
	
	\textcolor{blue} {/* Compute the index of the specific slice distributed to current thread for load and computation, respectively*/}\\
	\scriptsize
	$load\_index = lane\_id\ \%\ coalesce\_num$ \\
	$comp\_index = lane\_id\ /\ features.dim$ \\
	
	\textcolor{blue} {/* Compute the specific dimension of the coalescent features distributed to current thread for load and computation */}\\
	$comp_{dim} = lane\_id\ \%\ features.dim$ \\
	\textcolor{blue} {/* Check whether the current thread is distributed with work*/}\\
	
	\If {comp\_index $<$ coalesce\_num} {
		\scriptsize
		\textcolor{blue} {/*Compute the specific workload size (within a slice) for the thread*/}\\
		$load_{size}, comp_{size} = \textbf{compWork}(load\_index, comp\_index)$\\
		\textcolor{blue} {/*Load the target elements in the slice*/}\\
		\For{$nid \leftarrow lane\_id\ $ \KwTo $\ (slice\_size * coalesce\_num)$}{
			\textcolor{blue} {/*Compute the element offset in the slice for load*/}\\
			$load\_offset = nid\ /\ coalesce\_num$\\
			\If{$load\_offset > load_{size}$}{
				\textbf{break}
			}
			$\textbf{LoadSliceToSharedMem}(load\_offset)$ \\
			$nid\ +=\ coalesce\_num\ *\ features.dim$
		}
		\textcolor{blue} {/*Load the target dimension in the features and then compute*/}\\
		\For{$idx \leftarrow 0\ $ \KwTo $\ comp_{size}$}{
			\textcolor{blue} {/*Compute the element offset in the slice for computation*/}\\
			$comp\_offset\ =\ idx\ *\ coalesce\_num\ +\ comp\_index$ \\
			$Slice\ =\ \textbf{LoadSliceFromSharedMemToReg}(comp\_offset)$\\
			$Feature\ =\ \textbf{LoadFeatFromGlobalMemToReg}(comp_{dim})$\\
			$\textbf{MultiplyandACcumulate}(Slice, Feature, partial\_result)$\\
			$idx\ ++$
		}
		$\textbf{actomicAdd}(aggregation\_result, partial\_result)$
	} 
\end{algorithm}

\textbf{Thread-aware slice coalescing}. Even if merging multiple feature matrices to a coalescent one, the total row length (dimension) may still be less than 32 and the low thread utilization problem remains. In this situation, to further increase the number of active threads per warp, we coalesce adjacent slices in the adjacent matrix as a group and set it as the basic processing unit for each warp. Correspondingly, threads inside the same warp are uniformly divided into several thread groups (TGs) and each TG is assigned to exclusively operate on one slice. Then we implement the slice-group data layout on shared memory in a interleave pattern (\ding{184} in Figure \ref{fig:computation}) so that the access address of single data load is continuous from the warp view. Each TG operates on one element from the corresponding slice and one row from the coalescent feature matrices at each computing iteration. The maximal size of the slice group ($coalesce\_num$), namely the number of TGs per warp, is set as 4 to ensure each TG's data access granularity not exceed single memory transaction length (32 bytes). Algorithm \ref{alg:1} presents the detailed procedure of our parallel aggregation (for the small dimension cases) with the inputs: the coalescent features, the max number of nnz each slice owns ($32$) and $coalesce\_num$. In the implementation, each thread is first assigned to load elements in one particular slice from global memory to shared memory with a fixed stride and the slice index is directly corresponding to the thread index (Line 3 and 9-20). This design is based on the aforementioned interleaving data layout for each warp's continuous memory access. Then the thread fetches the data corresponding to one specific dimension of the features that is distributed to it (Line 6 and 26), loads the elements of one target slice on shared memory with the same stride pattern and finally performs the computation (Line 4 and Line 21-30).

\begin{figure*}[htbp]
	\centering
	\begin{minipage}[t]{.54\linewidth}
		\centering
		\includegraphics[scale=0.29]{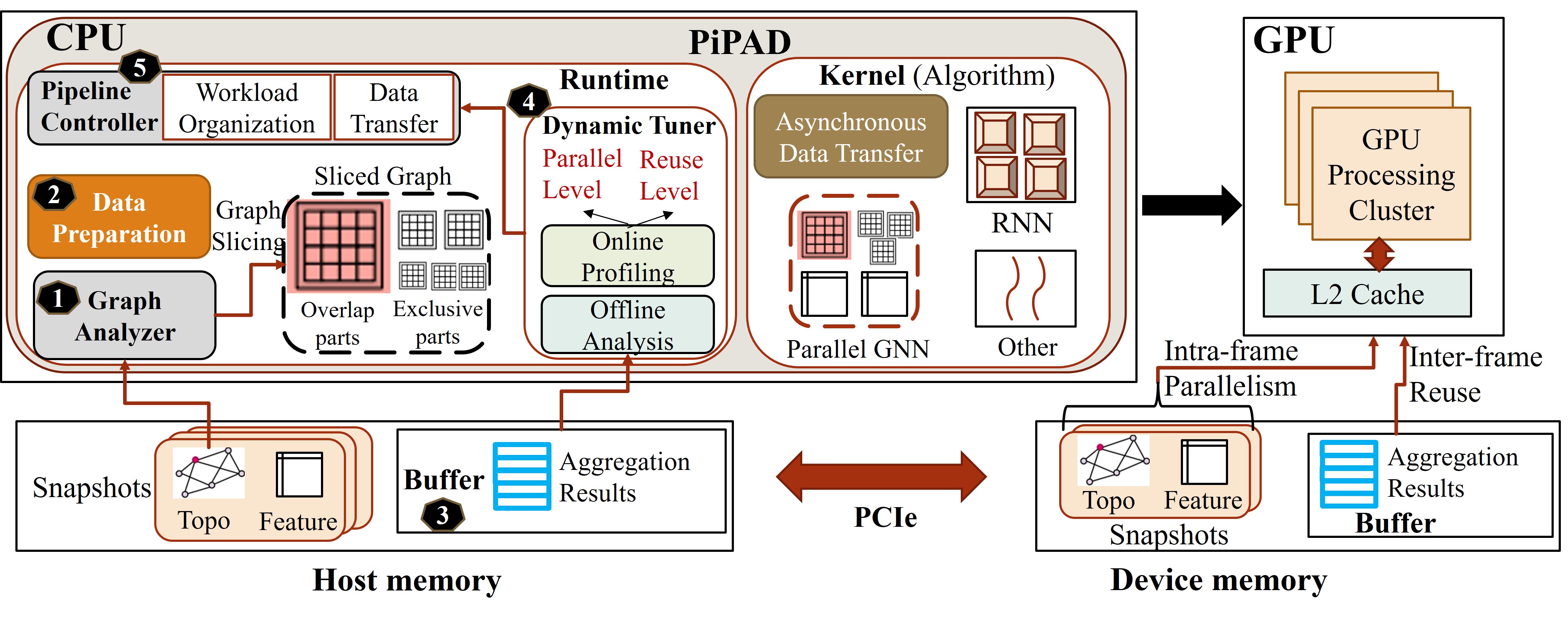}
		\caption{The Overall Architecture of PiPAD.}
		\label{fig:architecture}
	\end{minipage}
	\hspace{.1in}
	\begin{minipage}[t]{.4\linewidth}
		\centering
		\includegraphics[scale=0.29]{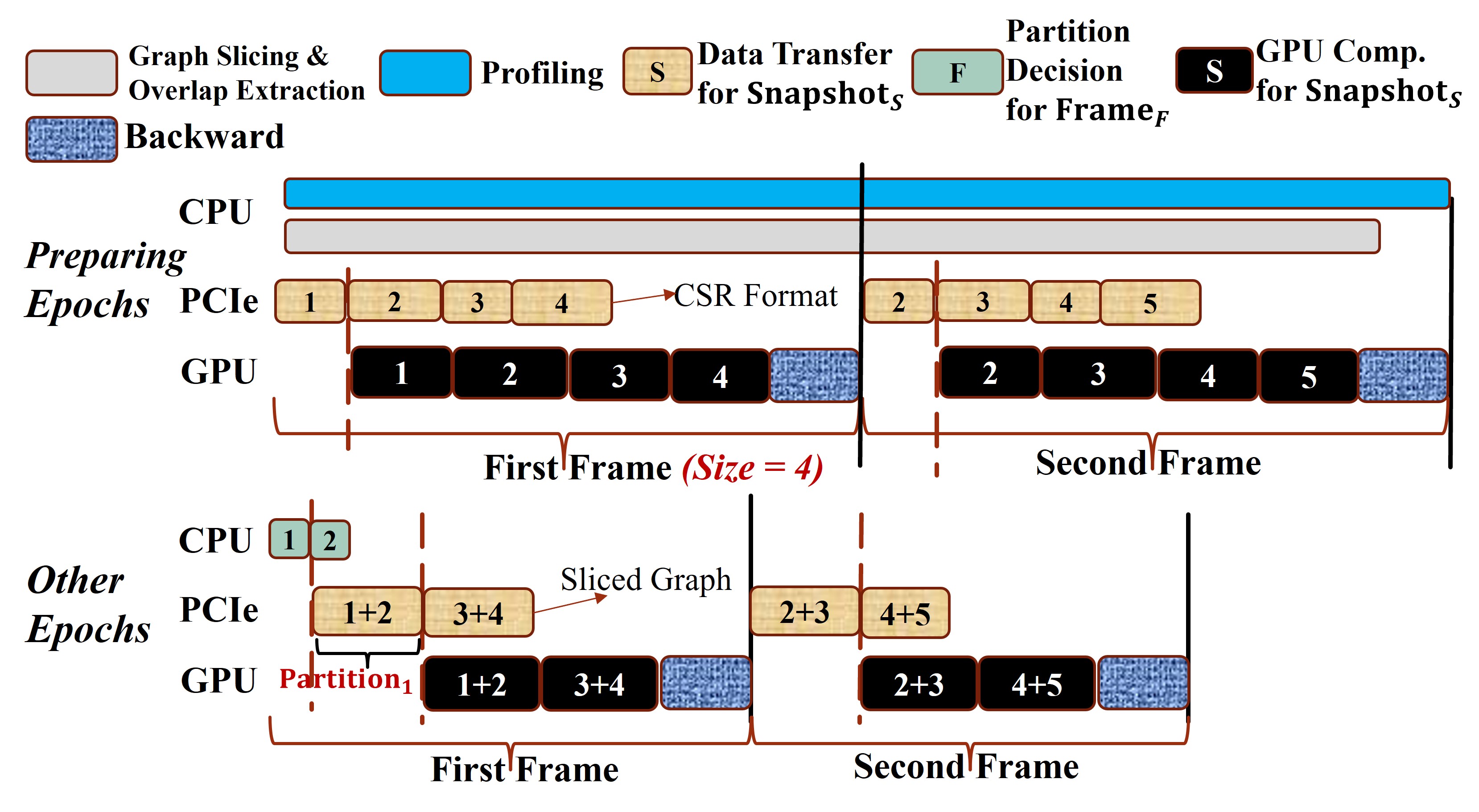}
		\caption{A Pipelined Execution Example.}
		\label{fig:pipeline}
	\end{minipage}
\end{figure*}

\textbf{Locality-optimized weight reuse}. To cooperate with the parallel aggregation, we utilize the weight reuse principle to enable the parallel update with better locality. Instead of naively conducting GEMM for one snapshot and then moving to the next, we choose to keep one weight tile staying on shared memory, and compute with the features of all snapshots successively (\ding{185}) and then move to the next weight tile for the same iteration. This locality-optimized computation paradigm can reuse the weights to the maximal extent. Note that this technique can not be applied to EvolveGCN since it updates the weights along the timeline.

For other kernels in DGNN, we execute them sequentially in the order of computation and data dependence. Besides, we reference OOB \cite{oh2022out} and leverage CUDA Graph API \cite{cudagraph} to launch these kernels together and reduce the issue overhead. 

\subsection{Pipeline Execution Framework}\label{sec:Pipe}
As illustrated in Figure \ref{fig:architecture}, PiPAD consists of the \textbf{algorithm-level} optimization for the efficient multi-snapshot computation and several \textbf{runtime-level} components to coordinate the general training process.

Specifically, \underline{in the algorithm level}, we devise the sliced CSR for convenient extraction of graph overlaps and reconstruct GNN to enable parallel computation over multiple snapshots within a frame. \underline{In the runtime level}, to fit with the new graph format, we first realize a low-overhead online \textit{graph analyzer} (\ding{182} in Figure \ref{fig:architecture}) to efficiently transform the format of all snapshots from the original CSR to ours. The adjacent matrices data for each partition includes the common overlap and exclusives for separate snapshots. A \textit{data preparation module} \ding{183} is used to conduct the overlap extraction among adjacent snapshots and prepare for the partition-wise computation manner. The above two components only need to function once during the first several epochs (namely \textit{preparing} epochs). In the following epochs, PiPAD performs the parallel training at the partition granularity for each frame. As for supporting data reuse among different frames, we maintain both \textit{CPU- and GPU-side buffer}s \ding{184} to cache the aggregation results. On the upper tier of our runtime, we implement a \textit{dynamic tuner} \ding{185} to online adjust the parallelism-level (the number of snapshots per partition inside a frame) and reuse-level (the size of aggregation results to cache in GPU-side), based on the offline analysis of our parallel GNN kernel and online profiling in the preparing epochs. Note that the preliminary offline experiments (\cref{sec:reuse}) only involve the GNN part for sensitivity analysis. Finally, our \textit{pipeline controller} \ding{186} regulates the overall training procedure.

Figure \ref{fig:pipeline} provides a PiPAD's pipelined execution example with the setting of two snapshots per partition and the frame size equal to 4. In the \textbf{preparing} epochs, the training abides by the traditional one-snapshot fashion with asynchronous data transfer. In the host, we collect the necessary statistics including the input data size, execution time and maximal memory consumption of each snapshot/frame. This profiling executes online to avoid early offline model-scale test and amortize the overhead for end-to-end performance improvements. These data will be leveraged to guide our dynamic tuner to choose the optimized parallel options without triggering the out-of-memory (OOM) exception (\cref{sec:reuse}). Besides, since extracting the graph overlaps and preparing the partition-wise adjacent matrices would also take time, we perform the graph slicing and subsequent extractions in the beginning once for all. The configurations of the snapshot amount per partition is a finite set (e.g., 2, 4 \& 8), thus the extraction overhead is controllable. Meanwhile, we also store the graph evolving rates among snapshots within each frame for the online tuning decision in the future. Then during \textbf{the following} epochs, the training proceeds in a partition manner with reduced data loading overhead and high computation parallelism. Compared to the canonical pattern, we first determine the parallelism-level for current frame (\cref{sec:reuse}), and then transport the partition-wise training data. Due to the performance similarity among different epochs in the training, we only perform this procedure once and stick to the generated configurations for each subsequent epoch. 

\textbf{Overhead analysis.} PiPAD introduces certain time and space overheads. First, our slice-based graph format requires extra space (\cref{sec:transfer}) and an auxiliary graph slicing process in the preparing epochs. We also need to extract the overlaps to build the overlap-based adjacent matrices for each partition previously. Note that our slicing strategy manages graphs at a fine granularity and can accelerate the extraction. By the actual measurements under our evaluation settings, those preprocessing can be accomplished within the first two epochs. Second, the data transfer for each partition can only proceed after the partition decision completes. The cost can also be amortized since this one-off operation is conducted asynchronously in CPU and overlap with other operations of the former partition/frame (Figure \ref{fig:pipeline}).

\subsection{Inter-frame Reuse and Dynamic Tuning}\label{sec:reuse}
This section presents our design of the data reuse across different frames and dynamic tuning on two key parameters.

\textbf{Inter-frame reuse}. As summarized in \cref{sec:motivation0}, the aggregation operation in the first GCN layer is unaffected by the parameter updating along the timeline. Therefore, a straightforward way is to store all relative aggregation results derived in the preparing epochs to CPU memory and reuse them for subsequent epochs. This can avoid repeatedly transmitting the adjacent matrices of those overlapped snapshots (for the models with only one single-layer GCN) and eliminate the redundant aggregation computation. But those aggregation results still need to be transferred to GPU for the next frame.

To further facilitate the reuse level, we maintain a GPU-side buffer to cache some aggregation results based on the used order in the next frame. The maximal buffer size is limited by the memory consumption of DGNN training process and GPU memory capacity. According to the frame-wise memory usage statistics provided by our profiling in the preparing epochs, we dynamically allocate the buffer with different sizes for each frame according to the demand. Note that, since the \textit{malloc/free} function execution is time-consuming, we only reallocate the buffer when its previous size is too small to accommodate the new intermediates.

\textbf{Dynamic Tuning}. To reduce the complexity of online decision-making, we uniformly distribute the snapshots in single frame to each partition. Then the dynamic tuner needs to determine two key parameters for each frame: the number of snapshots per partition ($S_{per}$) and the size of aggregation results to cache in GPU for the next frame reuse. The latter is decided by the memory usage conditions as explained above. With regard to $S_{per}$, there are three impact factors:

\begin{itemize}
	\item \textbf{Memory consumption}: Multiple-snapshot computing would increase the GPU memory usage at a certain stage compared to the one-snapshot pattern since processing each partition needs GPU to hold all related snapshots' data. We should avoid the OOM exception.
	\item \textbf{Computation speedup of the parallel GNN}: The acceleration effects of our intra-frame parallelism may vary over different input graphs and values of $S_{per}$.
	\item \textbf{Overlap level between the computation and data transfer}: Multi-snapshot processing not only promotes performance in the computing phase but also augments the data size of single transmission, which may stall the pipeline and lower GPU utilization instead.
\end{itemize}

\begin{figure}[htbp]
	\centering  
	\subfigure[Speedup of different $S_{per}$ settings as OR changes normalized to the one-snapshot execution.]{ 
		\centering    
		\includegraphics[scale=0.195]{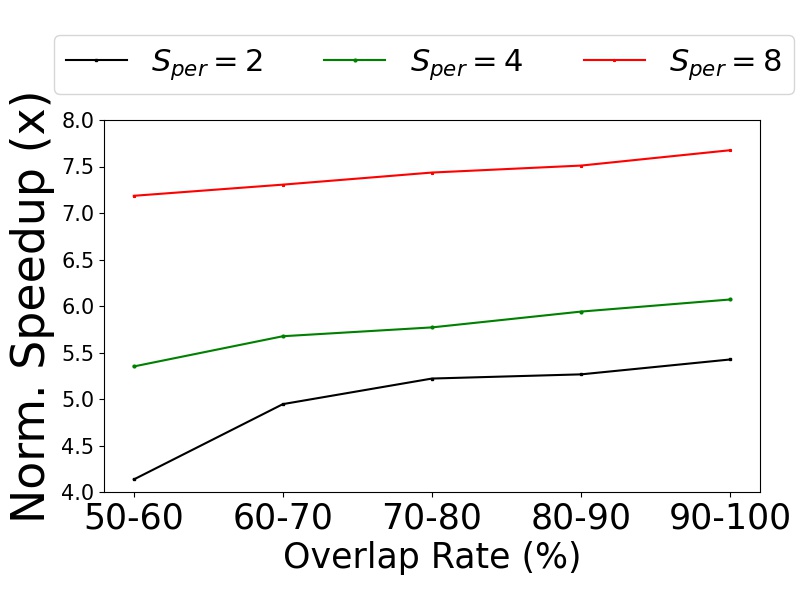}
		\label{fig:OR}
	}
	\subfigure[Normalized speedup of different $S_{per}$ settings as the feature dimension changes.]{   
		\centering    
		\includegraphics[scale=0.195]{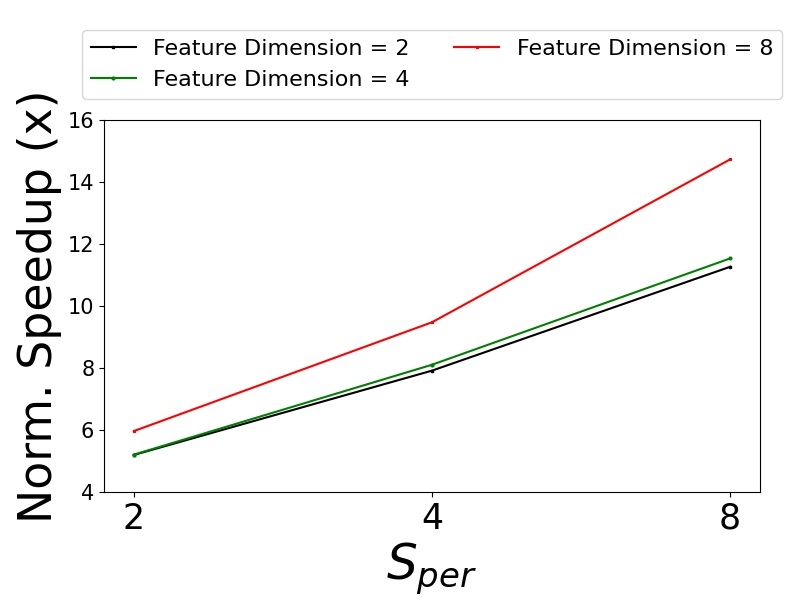}  
		\label{fig:fd}
	}
	\caption{The Offline Analysis Results of the Parallel GNN.}    
	\label{fig:offline}    
\end{figure}
We decide the value of $S_{per}$ with the following procedure.

First, since we employ the overlap-aware data organization method and the training data size gets reduced, the maximal memory consumption in the $N$-snapshot mode would not exceed $N$ times of that in the one-snapshot computation. Leveraging our online profiling statistics, we can derive an upper bound $U$ that assures our option not to trigger OOM.

Second, the computing speedup is mainly relevant to three elements: $S_{per}$, the node feature dimension of single graph and topology overlap rate (OR) among all snapshots within the partition. Simultaneously operating more snapshots normally means more parallelism but lower OR. Since the main optimization of our intra-frame parallelism lies in utilizing the overlaps to enable the efficient aggregation, the damage from low OR is not negligible. Therefore, we perform the offline analysis regarding our parallel GNN to guide the online decision. Figure \ref{fig:OR} and \ref{fig:fd} show the average speedup that different $S_{per}$ settings achieve across our datasets under various OR and feature dimension conditions, respectively. We construct the OR configurations with randomly selecting snapshot groups that satisfy the target overlap requirements. The results clearly show larger $S_{per}$ is preferred under the same OR or feature dimension settings. More importantly, the sketchy speedup distribution for separate OR sections can assist us to estimate the accelerating effects of different $S_{per}$ settings at runtime. We leverage the results from the offline analysis and the actual frame-wise topology changing rate statistics obtained online in the preparing epochs to derive the estimated execution latency of all available $S_{per}$ options.

Third, we previously analyze the data transfer latency for each option with the profiled data size information. Then, we reject those schemes rendering the pipeline stall and choose the one with the highest computing speedup from leftovers. With a specific parallelism and reuse option for each frame, our pipeline controller orchestrates the overall training.

\subsection{Discussion}\label{sec:implementation}
We implement PiPAD on top of PyTorch \cite{paszke2019pytorch} with C++, CUDA C and Python. But our methodology is general and independent of the specific deep learning or GNN frameworks.

\textbf{Limitation.} The multi-snapshot parallel training inevitably increases the phase-based memory consumption and causes the inherent scaling issue of large graphs. This limitation can be resolved through extending PiPAD to support multi-GPU training since our sliced CSR offers the convenience to further split the graphs. We leave it as our future work.

\section{Evaluation}
\subsection{Experiment Setup}\label{sec:setup}
\textbf{Models \& Datasets.} We choose three representative DGNN models (\cref{sec:dgnn}): MPNN-LSTM \cite{panagopoulos2021transfer}, T-GCN \cite{zhao2019t} and EvolveGCN \cite{pareja2020evolvegcn}. As shown in Table \ref{tab:dataset}, the datasets for our evaluation cover various real-world applications and are obtained from Network Repository \cite{rossi2015network, rossi2015networkrepository} (used by ESDG \cite{chakaravarthy2021efficient}), ASTGNN \cite{guo2021traffic} and MPNN-LSTM \cite{panagopoulos2021transfer}. Considering the memory capacity limit of single GPU, we set the input feature dimension of small and large-scale graph datasets to 16 and 2, respectively. And the hidden dimension is set to 32 and 6, respectively.
\begin{table}
	\caption{\label{tab:dataset}Graph Datasets for Evaluation. The number of vertices
		(\#N), edges (\#E), feature dimension (D) and edges after graph smoothening with edge-life \cite{chakaravarthy2021efficient} (\#E-S) across all snapshots as well as the number of snapshots (\#S) are shown.}
	\begin{center}
		\begin{tabular}{|c|c|c|c|c|c|}
			\hline
			\textbf{\footnotesize Dataset}&\textbf{\footnotesize\#N}&\textbf{\footnotesize\#E}&\textbf{\footnotesize D}&\textbf{\footnotesize\#S}&\textbf{\footnotesize\#E-S}\\
			\hline
			\multicolumn{6}{|c|}{\scriptsize \textit{Social Network} \cite{rossi2015network, rossi2015networkrepository}}\\
			\hline
			\textbf{\scriptsize Flickr} &\scriptsize 2.3 M &\scriptsize 33.1 M & \makecell[c]{\scriptsize 2}&  \makecell[c]{\scriptsize 132}&  \makecell[c]{\scriptsize 480 M} \\
			\hline
			\textbf{\scriptsize Youtube} &\scriptsize 3.2 M &\scriptsize 602 K & \makecell[c]{\scriptsize 2}&  \makecell[c]{\scriptsize 198}&  \makecell[c]{\scriptsize 11 M}\\
			\hline
			\multicolumn{6}{|c|}{\scriptsize \textit{E-commerce} \cite{rossi2015network, rossi2015networkrepository}}\\
			\hline
			\textbf{\scriptsize amz-Automotive} &\scriptsize 1.1 M &\scriptsize 1.3 M & \makecell[c]{\scriptsize 2}&  \makecell[c]{\scriptsize 524}&  \makecell[c]{\scriptsize 55 M}\\
			\hline
			\textbf{\scriptsize Epinions} &\scriptsize 727 K & \scriptsize 13.6 M & \makecell[c]{\scriptsize 2}&  \makecell[c]{\scriptsize 99}&  \makecell[c]{\scriptsize 78 M}\\
			\hline
			\multicolumn{6}{|c|}{\scriptsize \textit{Citation Network} \cite{rossi2015network, rossi2015networkrepository}}\\
			\hline
			\textbf{\scriptsize HepTh} & \scriptsize 22 K & \scriptsize 2.6 M & \makecell[c]{\scriptsize 16}&  \makecell[c]{\scriptsize 214}&  \makecell[c]{\scriptsize 18 M}\\
			\hline
			\multicolumn{6}{|c|}{\scriptsize \textit{Traffic Network} \cite{bai2020adaptive}}\\
			\hline
			\textbf{\scriptsize PEMS08} & \scriptsize 170 & \scriptsize 7202 & \makecell[c]{\scriptsize 16}&  \makecell[c]{\scriptsize 90}&  \makecell[c]{\scriptsize 7202}\\
			\hline
			\multicolumn{6}{|c|}{\scriptsize \textit{Disease Transmission} \cite{panagopoulos2021transfer}}\\
			\hline
			\textbf{\scriptsize Covid19-England} & \scriptsize 130 & \scriptsize 82 K & \makecell[c]{\scriptsize 16}&  \makecell[c]{\scriptsize 61}&  \makecell[c]{\scriptsize 108 K}\\
			\hline
		\end{tabular}
	\end{center}
\end{table}

\textbf{Baselines.} Since DynaGraph \cite{guan2022dynagraph} and CacheG \cite{li2021cache} restrict the input scenarios while ESDG \cite{chakaravarthy2021efficient} targets the distributed training, we compare PiPAD with the baseline \textbf{PyGT} and its three variants: (1) \textbf{PyGT-A}: an enhanced version of PyGT enabling asynchronous data transfer; (2) \textbf{PyGT-R}: a solution that integrates our inter-frame reuse mechanism into PyGT-A; (3) \textbf{PyGT-G}: compared to PyGT-R, PyGT-G replaces the original PyG-version of GCN module with GE-SpMM \cite{huang2020ge}, a state-of-the-art aggregation optimization work without previous node reordering over the graphs like \cite{wang2021gnnadvisor} and \cite{huang2021understanding}. GE-SpMM leverages shared memory to cache rows from sparse matrices and increases the locality for the CSR-based aggregation computation. We utilize this incremental comparison design to evaluate the overall performance and our individual optimization techniques simultaneously.

\textbf{Environment settings \& Metrics.} Our hardware platform is a 24-core Intel(R) Xeon(R) E5-2680 v3 CPU with 128GB host memory and one NVIDIA Tesla V100 GPU with 16GB HBM. We set the frame size to 16 and conduct all experiments on Ubuntu 18.04 with CUDA 11.3. And the version of PyGT and PyTorch is 0.53.0 and 1.10 respectively. In the overall performance comparison, we train for 200 epochs and measure the end-to-end training time including the graph slicing, data loading and GPU computation as well as the hardware utilization. We also perform the detailed analysis of our parallel GNN and sliced CSR. 

\subsection{Overall Performance Comparison}\label{sec:peformance}
\begin{figure}[htbp]
	\centering  
	\subfigure[EvolveGCN]{   
		\centering    
		\includegraphics[scale=0.15]{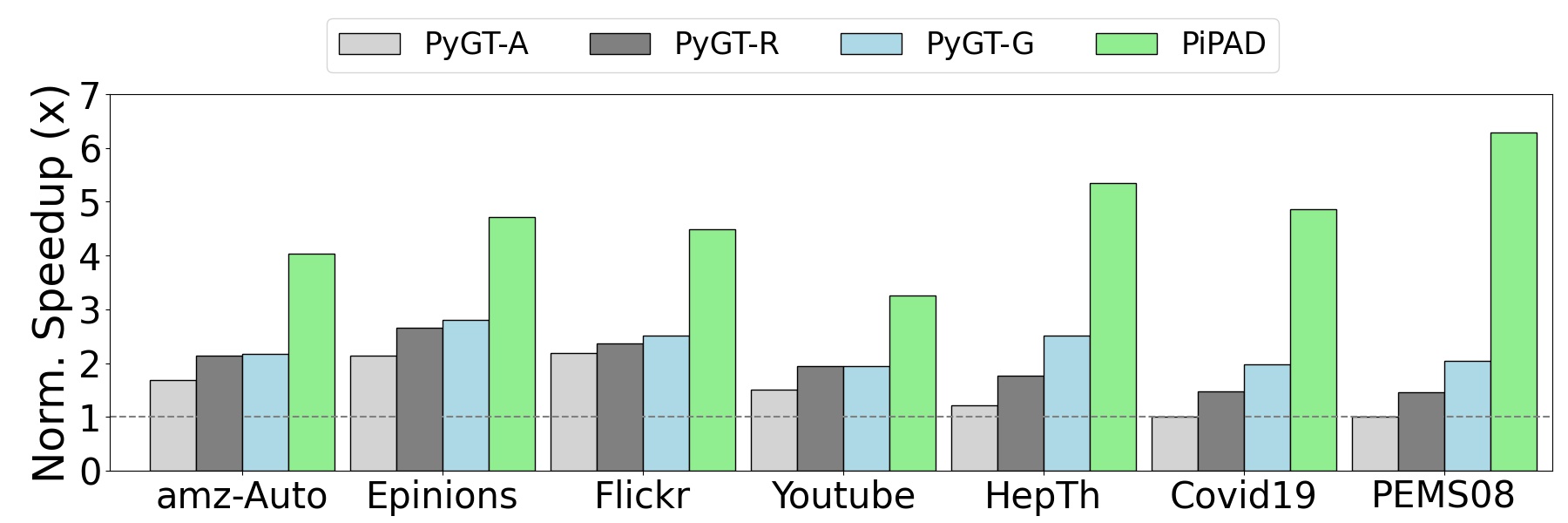}  
		\label{fig:perfEvolve}
	}
	\subfigure[MPNN-LSTM]{ 
		\centering    
		\includegraphics[scale=0.15]{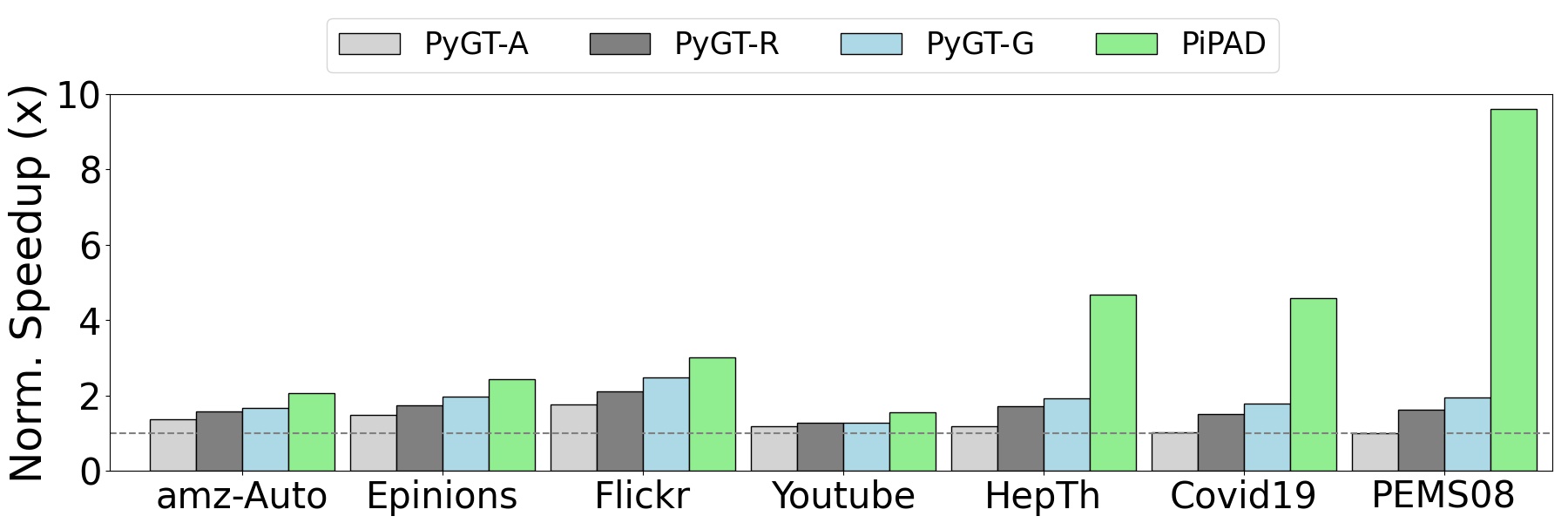}
		\label{fig:perfMPNN}
	}
	\subfigure[T-GCN]{   
		\centering    
		\includegraphics[scale=0.15]{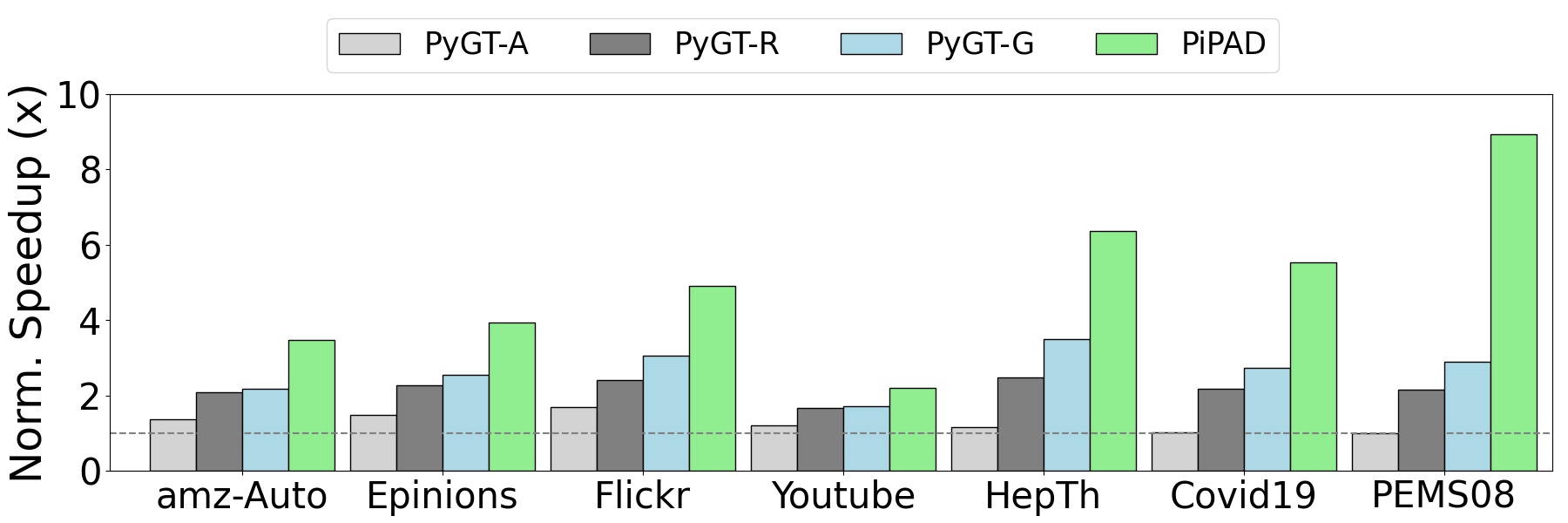}  
		\label{fig:perfTGCN}
	}
	\caption{Training Speedup over PyGT.}    
	\label{fig:overallperf}    
\end{figure}

Figure \ref{fig:overallperf} shows the end-to-end training time speedup of three models over PyGT. In general, with the holistic design covering data organization, transfer and computation manner, PiPAD outperforms all compared methods and achieves $1.54\times-9.57\times$ improvements over the baseline PyGT ($4.71\times$, $3.98\times$ and $5.18\times$ on average for EvolveGCN, MPNN-LSTM and T-GCN, respectively). As the small-scale datasets (HepTh, PEMS08 and Covid19) have less vertices and smaller feature vector size, our topology-wise data transfer reduction and computation optimizations can make a larger impact. Besides, due to the huge node amount and total memory usage, PiPAD can only enable 2-snapshot parallelism in the evaluation for the large datasets, which restricts our acceleration space. Thus our speedup is generally higher in the small-scale datasets while PyGT-A presents the opposite characteristic since the asynchronous transfer is its only optimization.

\begin{table*}[htbp]
	\centering
	\caption{GPU Utilization (\%) of Different Methods. AA: amz-Automotive; EP: Epinions; FL: Flickr; YT: Youtube; HT: HepTh; CE: Covid19-England; PE: PEMS08. The low values under small-scale datasets arise from the relatively larger CPU-side latency.}
	\scriptsize
	\begin{tabular}{|l||lllllll||lllllll||lllllll|}
		\hline
		\multirow{2}{*}{\fontsize{5.5pt}{\baselineskip}\selectfont Method} & \multicolumn{7}{c||}{EvolveGCN }                                                                                                                                      & \multicolumn{7}{c||}{MPNN-LSTM }                                                                                                                                  & \multicolumn{7}{c|}{T-GCN }                                                                                                                                   \\ \cline{2-22} 
		& \multicolumn{1}{l|}{AA}           & \multicolumn{1}{l|}{EP}     & \multicolumn{1}{l|}{FL}           & \multicolumn{1}{l|}{YT}            & \multicolumn{1}{l|}{HT}            &	\multicolumn{1}{l|}{CE}            &  PE           & \multicolumn{1}{l|}{AA}           & \multicolumn{1}{l|}{EP}     & \multicolumn{1}{l|}{FL}           & \multicolumn{1}{l|}{YT}            & \multicolumn{1}{l|}{HT}            &	\multicolumn{1}{l|}{CE}            &  PE           & \multicolumn{1}{l|}{AA}           & \multicolumn{1}{l|}{EP}     & \multicolumn{1}{l|}{FL}           & \multicolumn{1}{l|}{YT}            & \multicolumn{1}{l|}{HT}            &	\multicolumn{1}{l|}{CE}            &  PE           \\ \hline \hline
		{\fontsize{5.0pt}{\baselineskip}\selectfont PyGT}                           & \multicolumn{1}{l|}{\fontsize{5.0pt}{\baselineskip}\selectfont 75.18}          & \multicolumn{1}{l|}{\fontsize{5.0pt}{\baselineskip}\selectfont 70.75}         & \multicolumn{1}{l|}{\fontsize{5.0pt}{\baselineskip}\selectfont 76.47}         & \multicolumn{1}{l|}{\fontsize{5.0pt}{\baselineskip}\selectfont 78.85}          & \multicolumn{1}{l|}{\fontsize{5.0pt}{\baselineskip}\selectfont 31.9}         & \multicolumn{1}{l|}{\fontsize{5.0pt}{\baselineskip}\selectfont 12.04}          & \fontsize{5.5pt}{\baselineskip}\selectfont 10.87                 & \multicolumn{1}{l|}{\fontsize{5.0pt}{\baselineskip}\selectfont 84.81}          & \multicolumn{1}{l|}{\fontsize{5.0pt}{\baselineskip}\selectfont 80.44}         & \multicolumn{1}{l|}{\fontsize{5.0pt}{\baselineskip}\selectfont 81.36}         & \multicolumn{1}{l|}{\fontsize{5.0pt}{\baselineskip}\selectfont 89.91}          & \multicolumn{1}{l|}{\fontsize{5.0pt}{\baselineskip}\selectfont 36.01}         & \multicolumn{1}{l|}{\fontsize{5.0pt}{\baselineskip}\selectfont 11.89}          & \fontsize{5.0pt}{\baselineskip}\selectfont 10.99  & \multicolumn{1}{l|}{\fontsize{5.0pt}{\baselineskip}\selectfont 83.8}          & \multicolumn{1}{l|}{\fontsize{5.0pt}{\baselineskip}\selectfont 79.56}         & \multicolumn{1}{l|}{\fontsize{5.0pt}{\baselineskip}\selectfont 81.93}         & \multicolumn{1}{l|}{\fontsize{5.0pt}{\baselineskip}\selectfont 87.67}          & \multicolumn{1}{l|}{\fontsize{5.0pt}{\baselineskip}\selectfont 35.41}         & \multicolumn{1}{l|}{\fontsize{5.0pt}{\baselineskip}\selectfont 13.15}          & \fontsize{5.0pt}{\baselineskip}\selectfont 11.92        \\ \hline
		{\fontsize{5.0pt}{\baselineskip}\selectfont PyGT-A}                           & \multicolumn{1}{l|}{\fontsize{5.0pt}{\baselineskip}\selectfont 93.45}          & \multicolumn{1}{l|}{\fontsize{5.0pt}{\baselineskip}\selectfont 92.47}         & \multicolumn{1}{l|}{\fontsize{5.0pt}{\baselineskip}\selectfont 99.08}         & \multicolumn{1}{l|}{\fontsize{5.0pt}{\baselineskip}\selectfont 94.94}          & \multicolumn{1}{l|}{\fontsize{5.0pt}{\baselineskip}\selectfont 35.67}         & \multicolumn{1}{l|}{\fontsize{5.0pt}{\baselineskip}\selectfont 12.16}          & \fontsize{5.0pt}{\baselineskip}\selectfont 10.95                 & \multicolumn{1}{l|}{\fontsize{5.0pt}{\baselineskip}\selectfont 96.42}          & \multicolumn{1}{l|}{\fontsize{5.0pt}{\baselineskip}\selectfont 96.31}         & \multicolumn{1}{l|}{\fontsize{5.0pt}{\baselineskip}\selectfont \textbf{99.38}}         & \multicolumn{1}{l|}{\fontsize{5.0pt}{\baselineskip}\selectfont 97.94}          & \multicolumn{1}{l|}{\fontsize{5.0pt}{\baselineskip}\selectfont 39.14}         & \multicolumn{1}{l|}{\fontsize{5.0pt}{\baselineskip}\selectfont 11.92}          & \fontsize{5.0pt}{\baselineskip}\selectfont 11.0  & \multicolumn{1}{l|}{\fontsize{5.0pt}{\baselineskip}\selectfont 94.44}          & \multicolumn{1}{l|}{\fontsize{5.0pt}{\baselineskip}\selectfont 93.45}         & \multicolumn{1}{l|}{\fontsize{5.0pt}{\baselineskip}\selectfont 99.08}         & \multicolumn{1}{l|}{\fontsize{5.0pt}{\baselineskip}\selectfont 95.92}          & \multicolumn{1}{l|}{\fontsize{5.0pt}{\baselineskip}\selectfont 37.43}         & \multicolumn{1}{l|}{\fontsize{5.0pt}{\baselineskip}\selectfont 13.19}          & \fontsize{5.0pt}{\baselineskip}\selectfont 11.98         \\ \hline
		{\fontsize{5.0pt}{\baselineskip}\selectfont PyGT-R}                       & \multicolumn{1}{l|}{\fontsize{5.0pt}{\baselineskip}\selectfont \textbf{94.78}}          & \multicolumn{1}{l|}{\fontsize{5.0pt}{\baselineskip}\selectfont \textbf{94.45}}         & \multicolumn{1}{l|}{\fontsize{5.0pt}{\baselineskip}\selectfont \textbf{99.08}}         & \multicolumn{1}{l|}{\fontsize{5.0pt}{\baselineskip}\selectfont \textbf{96.03}}          & \multicolumn{1}{l|}{\fontsize{5.0pt}{\baselineskip}\selectfont \textbf{45.81}}         & \multicolumn{1}{l|}{\fontsize{5.0pt}{\baselineskip}\selectfont \textbf{13.6}}          & \fontsize{5.0pt}{\baselineskip}\selectfont \textbf{11.52}                 & \multicolumn{1}{l|}{\fontsize{5.0pt}{\baselineskip}\selectfont \textbf{97.3}}          & \multicolumn{1}{l|}{\fontsize{5.0pt}{\baselineskip}\selectfont \textbf{96.91}}         & \multicolumn{1}{l|}{\fontsize{5.0pt}{\baselineskip}\selectfont 99.16}         & \multicolumn{1}{l|}{\fontsize{5.0pt}{\baselineskip}\selectfont \textbf{98.65}}          & \multicolumn{1}{l|}{\fontsize{5.0pt}{\baselineskip}\selectfont 50.17}         & \multicolumn{1}{l|}{\fontsize{5.0pt}{\baselineskip}\selectfont \textbf{13.8}}          & \fontsize{5.0pt}{\baselineskip}\selectfont 12.51  & \multicolumn{1}{l|}{\fontsize{5.0pt}{\baselineskip}\selectfont \textbf{95.39}}          & \multicolumn{1}{l|}{\fontsize{5.0pt}{\baselineskip}\selectfont 94.5}         & \multicolumn{1}{l|}{\fontsize{5.0pt}{\baselineskip}\selectfont \textbf{99.13}}         & \multicolumn{1}{l|}{\fontsize{5.0pt}{\baselineskip}\selectfont 97.72}          & \multicolumn{1}{l|}{\fontsize{5.0pt}{\baselineskip}\selectfont \textbf{50.96}}         & \multicolumn{1}{l|}{\fontsize{5.0pt}{\baselineskip}\selectfont \textbf{15.36}}          & \fontsize{5.0pt}{\baselineskip}\selectfont 13.59          \\ \hline
		{\fontsize{5.0pt}{\baselineskip}\selectfont PyGT-G}                       & \multicolumn{1}{l|}{\fontsize{5.0pt}{\baselineskip}\selectfont 78.86}          & \multicolumn{1}{l|}{\fontsize{5.0pt}{\baselineskip}\selectfont 91.33}         & \multicolumn{1}{l|}{\fontsize{5.0pt}{\baselineskip}\selectfont 88.16}         & \multicolumn{1}{l|}{\fontsize{5.0pt}{\baselineskip}\selectfont 91.98}          & \multicolumn{1}{l|}{\fontsize{5.0pt}{\baselineskip}\selectfont 34.86}         & \multicolumn{1}{l|}{\fontsize{5.0pt}{\baselineskip}\selectfont 8.81}          & \fontsize{5.0pt}{\baselineskip}\selectfont 8.44                 & \multicolumn{1}{l|}{\fontsize{5.0pt}{\baselineskip}\selectfont 91.05}          & \multicolumn{1}{l|}{\fontsize{5.0pt}{\baselineskip}\selectfont 94.28}         & \multicolumn{1}{l|}{\fontsize{5.0pt}{\baselineskip}\selectfont 93.5}         & \multicolumn{1}{l|}{\fontsize{5.0pt}{\baselineskip}\selectfont 94.27}          & \multicolumn{1}{l|}{\fontsize{5.0pt}{\baselineskip}\selectfont 42.57}         & \multicolumn{1}{l|}{\fontsize{5.0pt}{\baselineskip}\selectfont 8.97}          & \fontsize{5.0pt}{\baselineskip}\selectfont 8.9  & \multicolumn{1}{l|}{\fontsize{5.0pt}{\baselineskip}\selectfont 89.5}          & \multicolumn{1}{l|}{\fontsize{5.0pt}{\baselineskip}\selectfont 93.55}         & \multicolumn{1}{l|}{\fontsize{5.0pt}{\baselineskip}\selectfont 92.25}         & \multicolumn{1}{l|}{\fontsize{5.0pt}{\baselineskip}\selectfont 93.46}          & \multicolumn{1}{l|}{\fontsize{5.0pt}{\baselineskip}\selectfont 40.64}         & \multicolumn{1}{l|}{\fontsize{5.0pt}{\baselineskip}\selectfont 10.47}          & \fontsize{5.0pt}{\baselineskip}\selectfont 11.23          \\ \hline
		{\fontsize{5.0pt}{\baselineskip}\selectfont PiPAD}                      & \multicolumn{1}{l|}{\fontsize{5.0pt}{\baselineskip}\selectfont 87.75}          & \multicolumn{1}{l|}{\fontsize{5.0pt}{\baselineskip}\selectfont 91.8}         & \multicolumn{1}{l|}{\fontsize{5.0pt}{\baselineskip}\selectfont 92.28}         & \multicolumn{1}{l|}{\fontsize{5.0pt}{\baselineskip}\selectfont 94.12}          & \multicolumn{1}{l|}{\fontsize{5.0pt}{\baselineskip}\selectfont 36.25}         & \multicolumn{1}{l|}{\fontsize{5.0pt}{\baselineskip}\selectfont 11.75}          & \fontsize{5.0pt}{\baselineskip}\selectfont 10.91                 & \multicolumn{1}{l|}{\fontsize{5.0pt}{\baselineskip}\selectfont 96.31}          & \multicolumn{1}{l|}{\fontsize{5.0pt}{\baselineskip}\selectfont 95.13}         & \multicolumn{1}{l|}{\fontsize{5.0pt}{\baselineskip}\selectfont 93.96}         & \multicolumn{1}{l|}{\fontsize{5.0pt}{\baselineskip}\selectfont 98.55}          & \multicolumn{1}{l|}{\fontsize{5.0pt}{\baselineskip}\selectfont \textbf{67.92}}         & \multicolumn{1}{l|}{\fontsize{5.0pt}{\baselineskip}\selectfont 12.6}          & \fontsize{5.0pt}{\baselineskip}\selectfont \textbf{14.25}  & \multicolumn{1}{l|}{\fontsize{5.0pt}{\baselineskip}\selectfont 94.39}          & \multicolumn{1}{l|}{\fontsize{5.0pt}{\baselineskip}\selectfont \textbf{94.68}}         & \multicolumn{1}{l|}{\fontsize{5.0pt}{\baselineskip}\selectfont 92.33}         & \multicolumn{1}{l|}{\fontsize{5.0pt}{\baselineskip}\selectfont \textbf{98.07}}          & \multicolumn{1}{l|}{\fontsize{5.0pt}{\baselineskip}\selectfont 50.61}         & \multicolumn{1}{l|}{\fontsize{5.0pt}{\baselineskip}\selectfont 12.18}          & \fontsize{5.0pt}{\baselineskip}\selectfont \textbf{14.23} \\ \hline
	\end{tabular}
	\label{tab:utilizationperf}
\end{table*}

PyGT-G obtains the second-best performance for almost all scenarios due to the three-fold optimizations: asynchronous data transfer to reduce transmission overhead, inter-frame reuse to eliminate the redundancy and GE-SpMM to accelerate the expensive aggregation. Compared with PyGT-G, our main advantage is the intra-frame parallelism that can not only process multiple snapshots for larger throughput but also facilitate the update function with the weight reuse. PyGT-R performs very close to PyGT-G or even better in some cases. First, employing inter-frame reuse would directly skip the aggregation in the first GCN layer and GE-SpMM targeting the aggregation acceleration thus turns nearly useless in T-GCN that executes multiple GCNs simultaneously and behaves like only owning one GCN. Second, GE-SpMM requires both CSR and Compressed Sparse Column (CSC) format to store the graphs for backward propagation in its implementation. This incurs more transfer overhead and damages the performance in large-scale datasets like Youtube.

The specific speedups vary greatly in different situations. We next perform the detailed analysis from the model aspect with the help of our experiments in Figure \ref{fig:breakdown} and Figure \ref{fig:GCNbreakdown}.

\textbf{EvolveGCN.} Incorporating two layers that each contain one GCN, EvolveGCN still needs to perform the aggregation in the GCN of the 2nd layer even if deployed with data reuse optimizations. This offers more acceleration space to our intra-frame parallelism. In addition, the other kernels besides GNN module in EvolveGCN only take small portions of total computation time (Figure \ref{fig:GCNbreakdown}), which further amplifies the effects of the parallel GNN. Therefore, PiPAD achieves the highest average speedup ($2.13\times$) against the second-best PyGT-G here among three models.

\textbf{MPNN-LSTM.} Contrast to other two models, MPNN-LSTM has the higher computation proportion and more time-consuming RNN components under the larger datasets with massive vertices (e.g., Youtube in Figure \ref{fig:breakdown} and \ref{fig:GCNbreakdown}). Therefore, when dealing with the large-scale datasets, PiPAD achieves the lowest speedup here among all cases (down to $1.22\times$ over PyGT-G). But the RNN-related execution latency of MPNN-LSTM is much smaller for small datasets (Figure \ref{fig:GCNbreakdown}). And MPNN-LSTM deploys a 2-layer GCN but dose not update weights along the timeline like EvolveGCN. The acceleration in both aggregation and update operation of our parallel GNN can fully function under those small-scale datasets (up to $9.57\times$ over PyGT).

\textbf{T-GCN.} With multiple GCNs executing in parallel, all aggregation operations are eliminated in T-GCN with the inter-frame reuse. Therefore, PyGT-R, PyGT-G and PiPAD all gain considerable speedups. But as mentioned above, the effects of GE-SpMM in PyGT-G are also removed with inter-frame reuse while PiPAD still can leverage the locality-optimized weight reuse to accelerate the update phase.

With PyTorch Profiler not support CUDA Graph, we utilize NVIDIA System Management Interface \cite{nvidiasystem} to collect the GPU utilization data where running memory copy kernels also counts towards the final utilization (Table \ref{tab:utilizationperf}). Therefore, PyGT-R and PyGT-A counter-intuitively perform better because PiPAD and PyGT-G greatly reduce the computation time (\cref{sec:detailed}) and amplify the percentage of total time taken by CPU-side operations. But we still outperform PyGT-G with the parallel GNN decreasing the number of issued kernels.
\subsection{Parallel GNN Analysis}\label{sec:detailed}
Since the data transfer greatly impacts the end-to-end training time especially in large-scale datasets, this section specially analyzes our algorithm-level optimization: intra-frame parallelism. We profile the execution time and kernel-level global memory access statistics of GNN (1-layer) module during the overall DGNN training process. PyGT and PyGT-G are chosen for comparison. We disable inter-frame reuse to thoroughly reveal the characteristics of GNN.
\begin{figure}[htbp]
	\centering  
	\subfigure[Speedup of GNN execution time (left-axis) and Memory access performance (right-axis)]{ 
		\centering    
		\includegraphics[scale=0.15]{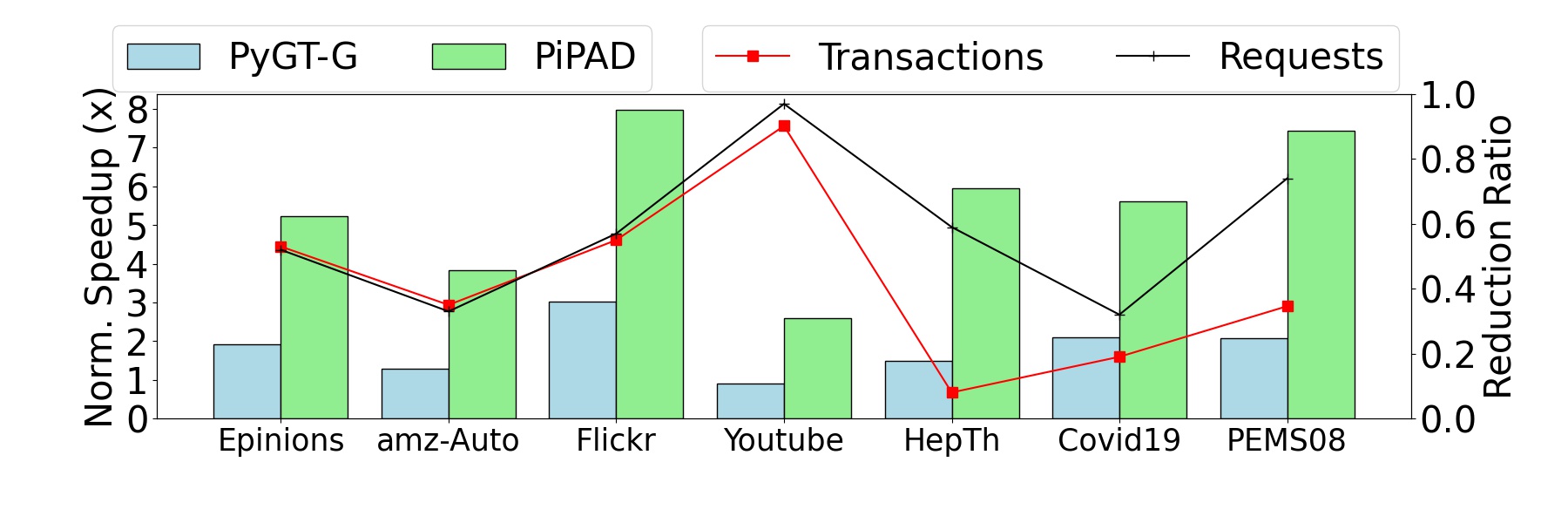}
		\label{fig:gnnperf}
	}
	\subfigure[Normalized speedup over PyGT as the feature dimension changes.]{   
		\centering    
		\includegraphics[scale=0.18]{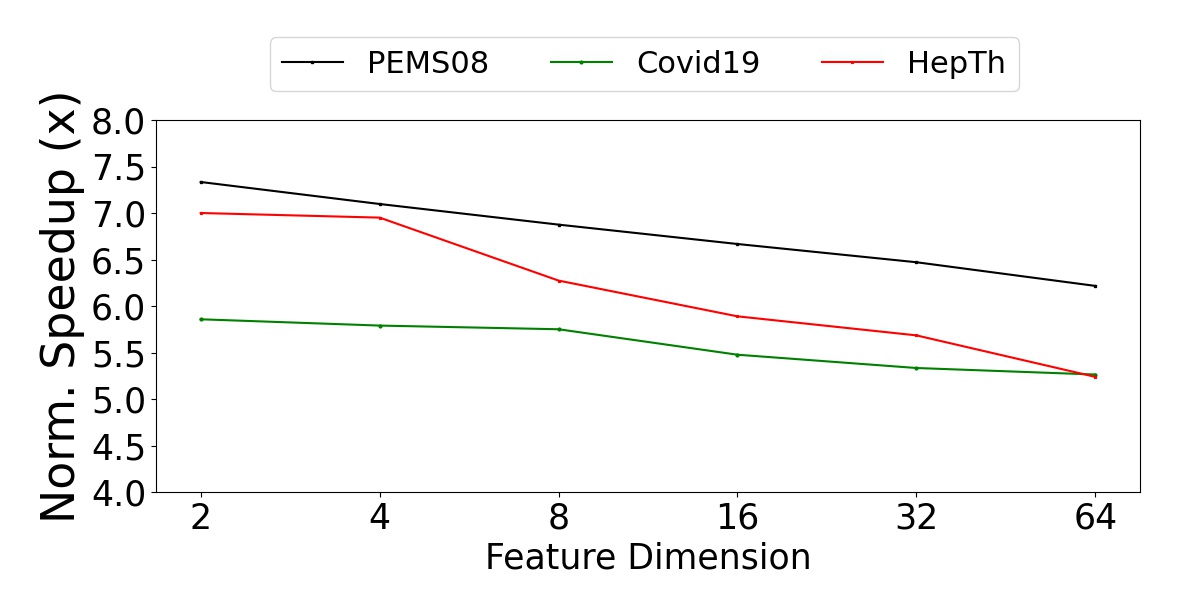}  
		\label{fig:sensitivity}
	}
	\caption{Detailed Analysis.}    
	\label{fig:detail}    
\end{figure}

\textbf{Speedup}. Figure \ref{fig:gnnperf} (left-axis) presents the GNN execution time speedup over PyGT. Compared to GE-SpMM following the inefficient one-snapshot manner and only targeting the SpMM-like aggregation, our parallel GNN processes multiple snapshots simultaneously and optimizes both aggregation and update phases. We achieve an average $5.6\times$ and $3.1\times$ improvements over PyGT and PyGT-G, respectively. From the dataset respective, since PiPAD and GE-SpMM both use shared memory to cache elements from the adjacent matrix in the aggregation, our higher speedups locate at the denser graph datasets with better locality in the graph structure (e.g., Flickr, HepTh and Epinions) similar to PyGT-G.

\textbf{Memory efficiency}. Unlike GE-SpMM only leveraging shared memory to optimize the access to the adjacent matrix, our parallel execution can enable coalescent memory access to multiple feature matrices. For the large datasets with 2-dimensional features, the parallel GNN can load the 16-byte feature with one transaction for two graphs simultaneously while GE-SpMM needs two individual transactions to do this. For the small-scale datasets with 16-dimensional features, single 64-byte data load generates one memory request and two transactions for one graph under GE-SpMM. But with the vector memory instructions, PiPAD can load the 256-byte data with one memory request and eight transactions for four graphs simultaneously. We access memory with the larger granularity while avoiding the \textit{request burst}. Hence, we can decrease the total number of global memory requests and transactions against PyGT-G by average $57\%$ and $45\%$ respectively (right-axis in Figure \ref{fig:gnnperf}). The improvements under Youtube stand out due to the excessive sparsity of this dataset. Numerous empty rows in CSR lead to the vast redundant memory accesses under GE-SpMM while our sliced CSR avoids this problem via the finer managing granularity.
\textbf{Thread utilization}. The warp\_execution\_efficiency metric measured via NVIDIA profiler \cite{nvidiaprofiler} directly shows the ratio of the average active threads per warp. To better demonstrate the low thread utilization issue, we set the input and hidden dimension of all seven datasets as 2 and 6. Our evaluation shows that the average thread utilization of the GNN-related kernels under PyGT-G is $57.2\%$ while PiPAD achieves $64.9\%$ based on our thread-aware slice coalescing.

\textbf{Sensitivity}. We perform the dimension sensitivity test to prove the dimension-awareness of our parallel GNN with the results shown in Figure \ref{fig:sensitivity}. Due to the memory capacity limit of our testbed GPU, we only test three small-scale datasets. Employing both vector memory instructions and thread-aware slice coalescing, our parallel GNN can achieve considerable speedups (at least $5.2\times$) for various feature dimension settings. But due to the memory consumption factor, we can enable the higher parallelism-level (larger $S_{per}$ in \cref{sec:reuse}) in the small-dimensional case for higher speedups.



\subsection{Sliced CSR Analysis}\label{sec:perfsliced}
\begin{figure}
	\centering
	\includegraphics[scale=0.18]{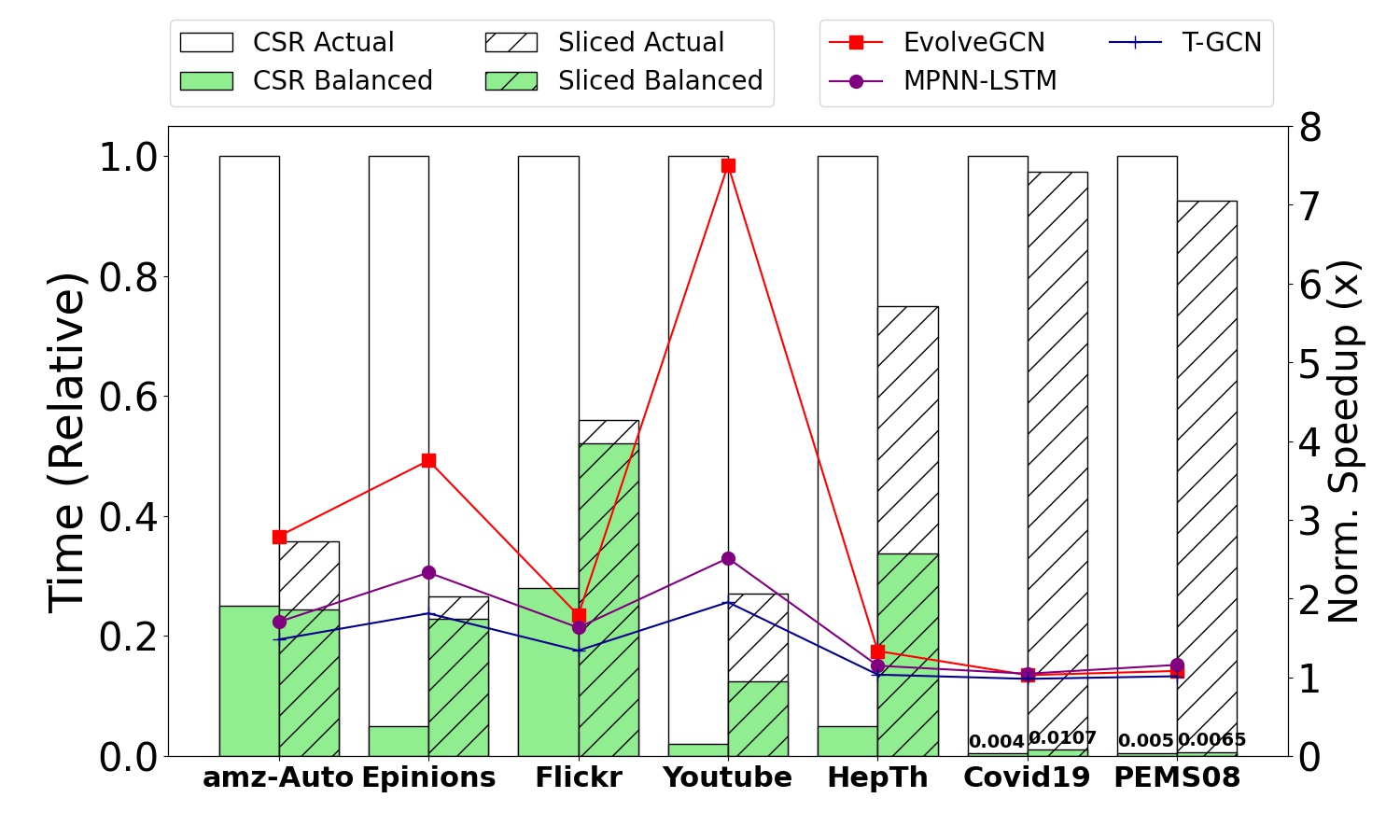}
	\caption{\label{fig:loadbalance} Load Balance Analysis and Overall Performance Comparison for the Sliced CSR.}
\end{figure}

We further implement the variant of PiPAD using the original CSR and conduct comparison experiments to analyze our sliced CSR. Referencing the methodology in \cite{huang2021understanding}, we plot our effects of promoting load balance for the GNN kernels in Figure \ref{fig:loadbalance} (left-axis). The Balanced bars (green-colored portion) represent the ideal execution latency in perfect load balance, derived through dividing the total time of all thread blocks by the maximal number of active thread blocks a GPU can accommodate \cite{huang2021understanding}. The gap between this ideal value and the actual execution time (Actual bars) reflects the degree of load imbalance, which is reduced by our sliced CSR across all datasets. The improvements are less significant under the small-scale datasets since they are more dense and easy to achieve load balance in the computation even with the original CSR. Besides, the speedups of overall training time from the model perspective (right-axis) show the similar trend to the results of load balance. Specially, the prominent improvement of EvolveGCN under Youtube is brought together by the large execution portion of GNN kernels in EvolveGCN (computation) and sliced CSR's less space usage than CSR due to the extreme sparsity of Youtube (data transfer).

\section{Conclusion}
This paper proposes PiPAD, a pipelined and parallel DGNN training framework to optimize the end-to-end performance on GPUs. With efficient parallel multi-snapshot processing and the runtime-level pipeline orchestration, PiPAD addresses the challenges of excessive data transmission, unexploited parallelism and memory access inefficiency in DGNN computing. Our evaluation shows that PiPAD achieves $1.22\times-9.57\times$ speedup over the state-of-the-art DGNN frameworks.
\bibliographystyle{ACM-Reference-Format}
\bibliography{reference}


\end{document}